%% file: main.tex
\documentclass[10pt,twocolumn,letterpaper]{article}

\newcommand{\figref}[1]{Fig.~\ref{#1}}
\newcommand{\tabref}[1]{Tab.~\ref{#1}}
\newcommand{\secref}[1]{Sec.~\ref{#1}}

\newcommand{\equref}[1]{Eq.~(\ref{#1})}

\newcommand{\methodName}{{DiffusionSeg}\xspace}

\usepackage{iccv}
\usepackage{times}
\usepackage{epsfig}
\usepackage{graphicx}
\usepackage{amsmath}
\usepackage{amssymb}
\usepackage{booktabs}
\usepackage{multirow}
\usepackage{bm}
\usepackage{float}
\usepackage{times}
\usepackage{xspace}
\usepackage{dsfont}
\usepackage{comment}
\usepackage{colortbl}
\usepackage{enumitem}
\usepackage{placeins}
\usepackage{footnote}
\usepackage{multicol}
\usepackage{graphbox}
\usepackage{mathrsfs} 
\usepackage{makecell}
\usepackage{threeparttable}
\usepackage{caption}
\usepackage{subcaption}
\usepackage{pifont}
\usepackage{listings}
\usepackage[normalem]{ulem}
\usepackage{marvosym}
\usepackage{soul}
\makeatletter 
  \newcommand\figcaption{\def\@captype{figure}\caption} 
  \newcommand\tabcaption{\def\@captype{table}\caption} 
\makeatother

\setstcolor{red}

\usepackage[pagebackref=true,breaklinks=true,letterpaper=true,colorlinks,bookmarks=false]{hyperref}

\iccvfinalcopy

\begin{document}

\title{DiffusionSeg: Adapting Diffusion Towards Unsupervised Object Discovery}

\author{Chaofan Ma$^{1*}$, \ Yuhuan Yang$^{1*}$, \ Chen Ju$^1$, \  Fei Zhang$^1$, \ Jinxiang Liu$^1$, \\
Yu Wang$^1$, \ Ya Zhang$^{1,2}$, \ Yanfeng Wang$^{1,2}\textsuperscript{\Letter}$\\
$^1$ Coop. Medianet Innovation Center, Shanghai Jiao Tong University \ \
$^2$ Shanghai AI Laboratory\\
{\tt\small \{chaofanma,\,yangyuhuan,\,ju\_chen,\,ferenas,\,jinxliu,\,yuwangsjtu,\,ya\_zhang,\,wangyanfeng\}@sjtu.edu.cn}
}

\twocolumn[{
\renewcommand\twocolumn[1][]{#1}
\maketitle
\begin{center}
    \centering
  	\captionsetup{type=figure}
  	\vspace{-0.6cm}
  	\includegraphics[width=1\linewidth]{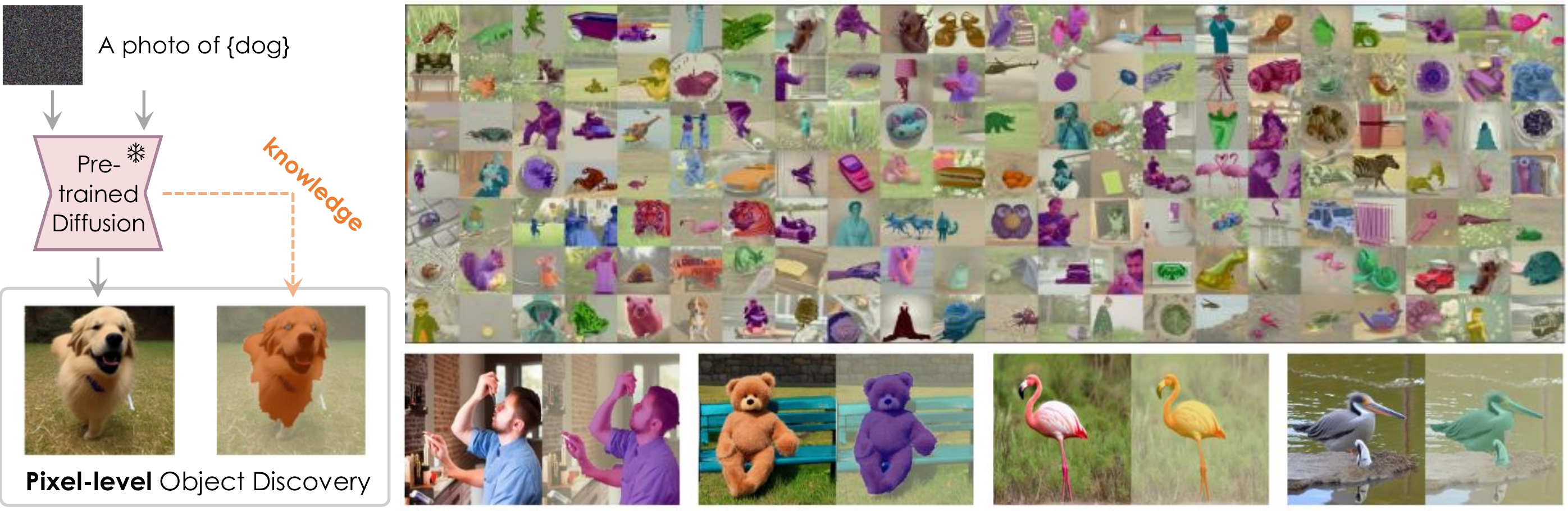} \\
    \vspace{-0.2cm}
    \captionof{figure}{\textbf{Left}: Insight for extracting pixel-level object masks by leveraging the visual knowledge from pre-trained text-to-image diffusion models. 
    \textbf{Right}: Qualitative visualization for extensive synthetic data and corresponding object masks. 
    }
	\label{fig:teaser}
\end{center}
}]

\maketitle

\input{sec/0-abstract}

\input{sec/1-introduction}

\input{sec/2-related}

\input{sec/3-methods}

\input{sec/4-experiments}

\input{sec/5-conclusion}

{\small
\bibliographystyle{ieee_fullname}
\bibliography{egbib}
}

\clearpage
\input{sec/6-appendix}

\end{document}

%% file: sec/0-abstract.tex
\begin{abstract}

Learning from a large corpus of data, pre-trained models have achieved impressive progress nowadays. As popular generative pre-training, diffusion models capture both low-level visual knowledge and high-level semantic relations. In this paper, we propose to exploit such knowledgeable diffusion models for mainstream discriminative tasks, \textit{i.e.}, unsupervised object discovery: saliency segmentation and object localization. However, the challenges exist as there is one structural difference between generative and discriminative models, which limits the direct use. Besides, the lack of explicitly labeled data significantly limits performance in unsupervised settings. To tackle these issues, we introduce \textbf{\methodName}, one novel synthesis-exploitation framework containing two-stage strategies. To alleviate data insufficiency, we synthesize abundant images, and propose a novel training-free AttentionCut to obtain masks in the first synthesis stage. In the second exploitation stage, to bridge the structural gap, we use the inversion technique, to map the given image back to diffusion features. These features can be directly used by downstream architectures. Extensive experiments and ablation studies demonstrate the superiority of adapting diffusion for unsupervised object discovery. 

\end{abstract}

%% file: sec/1-introduction.tex
\section{Introduction} 
\label{intro}

To date in the literature, large-scale pre-trained models, \ie, foundation models, have swept the CV 
domain for their remarkable progress. 
One general trend is pre-training then application, \ie, given a large corpus of data, first optimizes large-scale models to learn valuable prior knowledge about practical scenarios; then extracts some specific knowledge from pre-trained models for various downstream tasks.

Specifically, existing foundation models can be grouped into two branches, namely, discriminative (\eg, MoCo~\cite{moco}, DINO~\cite{dino}, CLIP~\cite{clip}) and generative (\eg, MAE~\cite{mae}, Diffusion~\cite{sohl2015deep, ddpm}). 
The two branches have their own advantages. 
\textit{Discriminative-based models} are trained to align images within the same class or with corresponding captions, thus they are aware of ``what'' the object is, \ie, better at \textit{high-level semantic tasks}, \eg, classification and retrieval. 
While \textit{generative-based models} are trained to capture both low-level visual knowledge (textures, edges, structures) and high-level semantic relations, thus they are aware of ``what'' and ``where'' the object is, \ie, better at \textit{pixel-level processing tasks}, \eg, reconstruction and segmentation. 
In terms of applications to downstream tasks, these two foundation models have large gaps. 
Discriminative-based models have been explored for both discriminative and generative tasks, \eg, detection~\cite{zareian2021open}, segmentation~\cite{ma2022open}, image synthesis~\cite{wang2022clip} and video understanding~\cite{ju2022prompting,ju2022distilling}. However, since discriminative pre-training focuses more on high-level semantics, it is difficult to deal with dense prediction tasks well. While generative pre-training, with both low-level and high-level visual knowledge, is now still stuck in the limited applications of low-level tasks, \eg, image generation~\cite{nichol2021glide}, colorization~\cite{saharia2022palette}, visual inpainting~\cite{esser2021imagebart}.

Hence, a novel question naturally raises: \textit{is generative-based pre-training also or even more valuable for the mainstream discriminative tasks?}
This paper makes a step towards positively answering the question, \ie, we adopt popular diffusion models to solve object discovery, \ie, saliency segmentation and object localization. The \textit{unsupervised} setting is explored to clearly evaluate the effectiveness.

To adapt pre-trained diffusion models for downstream tasks, a vanilla idea is to directly use the feature inside the model. However, it is infeasible, as there are considerable gaps lying across diffusion models and discriminative object discovery. 
(1) The structural difference between generative and discriminative models limits the direct transfer, \ie, diffusion turns noise into random images, while object discovery finds masks from given images. 
(2) Lacking explicitly labeled data significantly limits training performance of downstream tasks, especially for unsupervision.

In this paper, we design one novel synthesis-exploitation framework, containing two-stage strategies to tackle the above two issues respectively.
Specifically, the first synthesis stage is designed to tackle the issue of insufficient labeled data. We propose novel training-free  AttentionCut to obtain masks during synthesizing sufficient images. 
Images are synthesized using text-to-image diffusion model with random noise and category as inputs. 
Masks are generated leveraging cross- and self- attention in this diffusion model. 
As shown in Fig.~\ref{fig:teaser}, these synthetic images are realistic with accurate mask, which is impressive and demonstrates the quality. 
The second exploitation stage is proposed to bridge the structural gap. We combine inversion technique with diffusion models, to deterministically map the given image back to diffusion features. 
This allows the diffusion model to be regarded as a universal knowledge extractor, which can be directly used by any downstream architecture.
Results show the strong capabilities of this knowledge and training a lightweight decoder can unify the utilization of diffusion pre-training and object discovery.

On six standard benchmarks, namely, ECSSD, DUTS, DUT-OMRON for segmentation, while VOC07, VOC12, COCO20K for detection, our method significntly outperforms existing state-of-the-art methods. We also conduct extensive ablation studies to reveal the effectiveness of each component, both quantitatively and qualitatively. 

To sum up, our contributions lie three fold:

$\bullet$ We pioneer the early exploration in adapting free pixel-level knowledge from pre-trained diffusion models to facilitate unsupervised object discovery;

\vspace{0.05cm}
$\bullet$ We design a novel synthesis-exploitation framework that  explicitly extracts knowledge through data synthesis and leverages implicit knowledge by diffusion inversion;

\vspace{0.05cm}
$\bullet$ We conduct extensive experiments and ablations to reveal the significance of adapting diffusion knowledge and our superior performance on six public benchmarks.

%% file: sec/2-related.tex
\section{Related Work}

{\noindent \bf Generative Models} can roughly be classified into two main branches: GANs and diffusions. 
As the early representatives, GANs~\cite{gan, mirza2014conditional,isola2017image,zhu2017unpaired,karras2019style,brock2018large,karras2020analyzing} have the advantage of generating realistic and diverse data that are similar to the original. They can also learn complex and high-dimensional distributions without explicit density estimation. Such properties allow GANs to enjoy success in image generation~\cite{karras2019style}, image-to-image translation~\cite{zhu2017unpaired,huang2018multimodal,isola2017image}. However, they are hard to train stably. Without careful tuning of hyperparameters, they usually suffer from mode collapse, \ie, only generating a few modes of data distribution. 
In contrast, diffusion models~\cite{sohl2015deep,ddpm,song2019generative,song2020score,dhariwal2021diffusion,nichol2021improved, ddim} have recently broken the long-term dominance of GANs and raised the bar for generative modeling. Compared with GANs, they are friendly for using, without need for adversarial training. Besides, they also achieve state-of-the-art image quality and fidelity on various datasets~\cite{dhariwal2021diffusion}.
Benefited from such advantages, diffusion models have been applied to various generative tasks, such as text-to-image generation~\cite{nichol2021glide, ramesh2022hierarchical, saharia2022photorealistic, ldm}, colorization~\cite{saharia2022palette}, super-resolution~\cite{saharia2022image}, inpainting~\cite{esser2021imagebart, lugmayr2022repaint}, and semantic editing~\cite{choi2021ilvr, meng2021sdedit}.

Nevertheless, all above methods focus on preliminary generation tasks. In this paper, we explore the significance of generative pre-training models for discriminative tasks. The insight is that generative models are pre-trained to contain both low-level knowledge and high-level semantic relations.
Among all generative models, we choose diffusion models as representatives, for their impressive performance.

\begin{figure*}[t]
    \centering
    \vspace{-0.2cm}
    \includegraphics[width=0.97\textwidth]{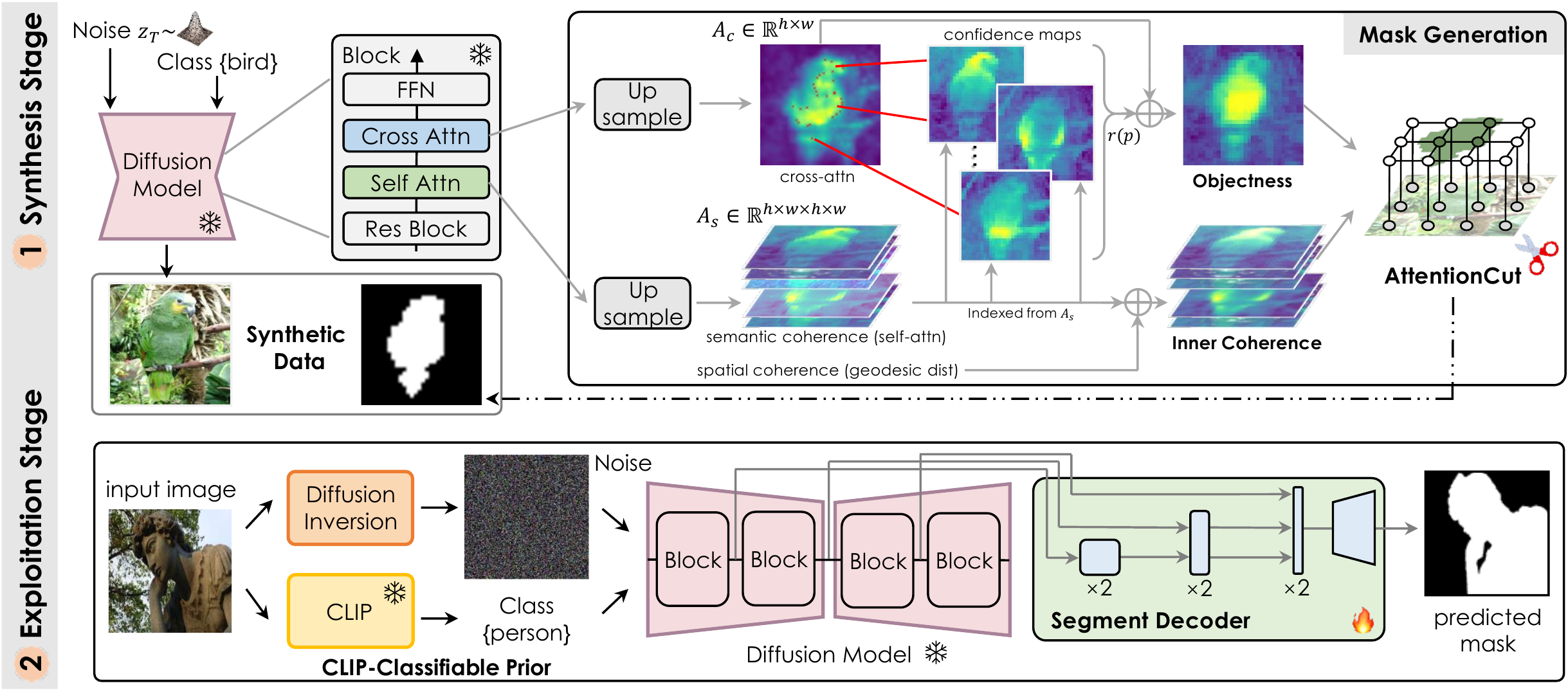}
    \vspace{-0.2cm}
    \caption{We use pre-trained diffusion models to synthesize extensive data, helping the model training, then evaluated on real images. 
    \textbf{(1) The Synthesis Stage.} We synthesize free image-mask pairs, where mask generation is solved leveraging cross- and self- attention by AttentionCut.
    \textbf{(2) The Exploitation Stage.} We extract knowledge 
    using diffusion inversion, then only one lightweight decoder is trained for object discovery on synthetic data.
    } 
    \label{fig:framework}
\end{figure*}

\vspace{0.2cm}
{\noindent \bf Object Discovery} aims at detecting and segmenting salient objects in the natural scenes, consisting of two popular sub-tasks: saliency segmentation and object localization. 
Existing methods have two settings: supervised and unsupervised.
The supervised methods~\cite{hou2016deeply, zhao2019pyramid, qin2020u2, yun2022selfreformer} are trained with large-scale pixel-level human annotations, which are time-consuming and expensive to acquire.

By contrast, the unsupervised setting without any labor labels, has received increasing attentions. Concretely, most methods embrace discriminative-based pre-trained models for help. 
LOST~\cite{lost}, Deep Spectral~\cite{MelasKyriazi2022DeepSM}, and TokenCut~\cite{tokencut} leverage features from self-supervised ViTs~\cite{dino} with contrastive learning~\cite{ouali2020autoregressive, ji2019invariant, ju2023constraint, ju2022adaptive} that exhibit object segmentation potential, after which a heuristic strategy or a graph-based method~\cite{shi2000normalized} is employed. SelfMask~\cite{selfmask} revisits the spectral clustering on image features from various self-supervised models, \eg, MoCo~\cite{moco}, SwAV~\cite{caron2020unsupervised}, DINO~\cite{dino}, to obtain pseudo-labels, which are then used to train one salient object detector.
FreeSOLO~\cite{Wang2022FreeSOLOLT} proposes to generate correlation maps which are then ranked and filtered by maskness scores. 
DINOSAUR~\cite{seitzer2022bridging} reconstructs features from self-supervised models for object-centric representations. Different with all above methods only using discriminative pre-training, this paper proves that generative-based pre-training is also or even more valuable for mainstream discriminative tasks, by a synthesis-exploitation strategy.

%% file: sec/3-methods.tex
\section{Methods}

This paper aims to utilize pre-trained diffusion generation models for downstream tasks by proposing a two-stage synthesis-exploitation framework. 
In \secref{Problem Formulation}, we start by describing the preliminary. 
In \secref{Synthesizing Labeled Data}, we detail the synthesis stage to generate sufficient labeled data. 
In \secref{Diffusion Features and Synthetic Data Supervision}, we detail the exploitation stage to close the structural gap between generative models and discriminative tasks.

\subsection{Preliminary and Overview}
\label{Problem Formulation}
{\noindent \bf Problem Definition.} Object Discovery (OD), \ie, saliency segmentation and object localization, as a fundamental and typical discriminative task, is studied in this paper. 
Concretely, object discovery aims to train one pixel-level segmentation model $\Phi_{\mathrm{OD}}$ that partitions one image $\mathcal{I}$ into two disjoint groups, namely, foreground and background.
\vspace{-0.5em}
\begin{equation}
    \mathcal{M}_{\mathrm{seg}} = \Phi_{\mathrm{OD}}(\mathcal{I}) \in\{0,1\}^{H \times W \times 1}, \  \mathcal{I} \in \mathbb{R}^{H \times W \times 3}, 
\vspace{-0.5em}
\end{equation}
where $\mathcal{M}_{\mathrm{seg}}$ refers to the binary segmentation mask. 

Here, to clearly evaluate the effectiveness of our method, we focus on the strict \textit{unsupervised} setting, {\em i.e.}, the model is trained \textit{without} any manually annotated data.

\vspace{0.1cm}
{\noindent \bf Motivation.} 
\hspace{1pt} This paper aims to exploit pixel-level visual knowledge from pre-trained diffusion generation models, for downstream discriminative tasks, \eg, OD. To achieve this goal, we design a novel synthesis-exploitation framework (\figref{fig:framework}). Specifically, at the synthesis stage, we explicitly construct one free (infinite-size) discriminative synthetic dataset, to obtain sufficient labeled samples. At the exploitation stage, we enable diffusion to be compatible with OD tasks, by extracting implicit diffusion features, and training one discovery decoder with the synthetic dataset.

\vspace{0.2cm}
{\noindent \bf Diffusion}~\cite{sohl2015deep,ddpm} is one recently popular generative idea, containing forward and reverse processes. 
The \textit{forward process} is a Markov chain where noise is gradually added to the data.
The \textit{reverse process} is a denoising procedure that can be decomposed into a linear combination of a noisy image $\boldsymbol{x}_t$ and a noise approximator $\epsilon_\theta(\cdot)$. $t=1,\dots,T$ refers to the denoising timesteps.
The key to diffusion models is to learn the function $\epsilon_\theta(\cdot)$, typically using a UNet~\cite{ronneberger2015u}.

Particularly, we build on a variant of the text-to-image diffusion model, namely, Stable Diffusion~\cite{ldm}. During the synthesis process, it's sampled by iteratively denoising $\boldsymbol{x}_t$ conditioned on the input text prompt $y$ for timestep $t=1,\dots,T$. The conditional denoising UNet $\boldsymbol\epsilon_\theta(\boldsymbol{x}_t, t, y)$ stacks layers of self- and cross-attentions. 
$y$ is first encoded to text embeddings by a pre-trained text encoder, then text embeddings are mapped to intermediate layers as $K$ and $V$ via the attention mechanism, and the noisy image $\boldsymbol{x}_t$ is mapped as $Q$. 
For step $t$ and layer $l$, we call cross-attention as $\mathcal{A}_c^{t,l}$, self-attention as $\mathcal{A}_s^{t,l}$, and intermediate features as $\mathcal{F}^{t,l}$.
{\bf Note that}, this paper freezes Stable Diffusion pre-trained on LAION-5B~\cite{schuhmann2022laion} (5 billion image-text pairs), as a knowledge provider. This diffusion model involves both low-level object details and high-level class semantics, enabling us to achieve unsupervised object discovery.

\subsection{Synthesis Stage:  Free Data Generation}
\label{Synthesizing Labeled Data}
As illustrated in \figref{fig:framework} (1), this stage aims to synthesize large and free image-mask pairs through Stable Diffusion, solving the lack of labeled training data under unsupervised settings. 
We detail image synthesis in \secref{sec:Image Generation}, and mask generation in \secref{synthetic mask generation}.

\vspace{-0.25cm}
\subsubsection{Image Generation}
\label{sec:Image Generation}
For one pre-trained text-to-image Stable Diffusion~\cite{ldm}, we here freeze it, then generate images through inputting random Gaussian noise and class text prompts. Class names are sampled from ImageNet~\cite{imagenet}.

For text input, a simple way is to simply use class names, but this may limit diversity and cause bottlenecks for downstream tasks. 
Hence, to adaptively generate various text prompts for each class, we interact with ChatGPT~\cite{chatGPT}. 
For example, we ask ChatGPT to list prompts about ``aeroplane'', then it could give some generative-style prompts like: ``\textit{A aeroplane soaring through a vibrant sunset sky, fluffy clouds, warm lighting, viewed from a low angle, realistic style.}'' 
The generated prompts introduce richer context, thus can better unleash the potential of the Stable Diffusion to synthesis high-fidelity, more diverse images. One noise reduction strategy is also applied following~\cite{he2022synthetic}.

\vspace{-0.25cm}
\subsubsection{Mask Generation}
\label{synthetic mask generation}

Here, we generate high-quality masks by leveraging attentions in pre-trained diffusion models as clues, following two non-trivial observations.
(1) Cross-attention $\mathcal{A}_c$ indicates locality between the conditioning text and noisy image, thus $\mathcal{A}_c$ can coarsely describe \textit{objectness}.
(2) Self-attention $\mathcal{A}_s$ inside one image indicates pairwise semantic similarity between pixels, thus $\mathcal{A}_s$ could roughly describe \textit{coherence}.
Inspired by these, we propose AttentionCut, a training-free strategy to generate masks guided by attention maps.

\vspace{0.2cm}
{\noindent \bf Preparations.}
We first extract $\mathcal{A}_c$ and $\mathcal{A}_s$ at the position of category token in the prompt sentence, then aggregate different resolutions and timesteps considering multi-scale objects and avoiding focus shift during diffusion.
Formally,
\vspace{-0.5em}
\begin{equation}
    \mathcal{A}_c=\frac{1}{kT}\sum_{l=1}^{k}\sum_{t=0}^{T-1} \mathcal{A}_c^{t,l}; \
    \mathcal{A}_s=\frac{1}{LT}\sum_{l=1}^{L}\sum_{t=0}^{T-1} \mathcal{A}_s^{t,l}, 
    \vspace{-0.5em}
\end{equation}
where $t=T-1, \dots, 0$ is for each reverse step and $l = 1, \dots, L$ is for intermediate layers. $\mathcal{A}_c$ is averaged among the top-$k$ of the standard variation from all $\mathcal{A}_c^l$, while $\mathcal{A}_s$ is averaged among all layers and time steps.

\vspace{0.2cm}
{\noindent \bf Objectness.}
Intuitively, the pixel-level cross-attention $\mathcal{A}_c$ under a specific category can roughly be seen as segmentation masks, as it indicates how likely a pixel belongs to the category. However, in practice we found $\mathcal{A}_c$ is sparse and inattentive near the boundary, which can seriously damage segmentation results. 
To handle this issue, we improve $\mathcal{A}_c$ by strengthening the edge area with the self-attention $\mathcal{A}_s$. It indicates semantic connectivity, \ie, how semantically two pixels belong to one group. 
Specifically, we first randomly select a set of initial seeds $\mathcal{B}$ from the boundary of the binary mask $\left[\mathcal{A}_c>\tau\right]$. 
Then each selected seed $b \in \mathcal{B}$ can expand as a confidence map $\mathcal{A}_s(b, \cdot)$, which is the self-attention between $b$ and other pixels, indicating weights of the boundary area. 
We assume $\mathcal{A}_s(\cdot, b)=\mathcal{A}_s(b, \cdot)$, as $\mathcal{A}_s$ is symmetric theoretically.
For pixel $p$, these maps are averaged as a refined map $r(p)$, to reinforce the boundary pixels: 
\vspace{-0.5em}
\begin{equation}
    r(p) = 1/{|\mathcal{B}|}\cdot \sum\nolimits_{b \in {B}} \mathcal{A}_s(p,b).
\vspace{-0.5em}
\end{equation}

Combining cross-attention $\mathcal{A}_c$ and the refined map $r(p)$ with a balance weight $\lambda_\phi$, the pixel-level objectness $\phi$ are:
\vspace{-0.5em}
\begin{equation}
    \phi(p) = \left\{
    \begin{aligned}
        -\log(\mathcal{A}_c(p)+\lambda_\phi r(p)),\  \text{if }p \in\text{foreground},\\
        \log(1-\mathcal{A}_c(p)-\lambda_\phi r(p)),\  \text{if }p\in\text{background},
    \end{aligned}
    \right.
    \vspace{-0.5em}
\end{equation}
where $\mathcal{A}_c(p)$ is the cross-attention at pixel $p$.

\vspace{0.2cm}
{\noindent \bf Inner Coherence.}
With only objectness, we found that the masks tend to lose local information, for example, irregular corners, mis-segmented holes, or jagged contours.
This can be solved by taking local consistency into account, \ie, how likely two neighboring pixels belong to one group.
Here we design an inner coherence term that can help to enforce continuity, proximity and smoothness of segments belonging to the same object, and penalize those who deviate.

The proposed inner coherence consists of two parts: semantic and spatial.
As mentioned above, $\mathcal{A}_s$ can indicate semantic coherence, as self-attention is calculated in semantic feature space.
Spatial coherence is designed to indicate pixels pairwise distance in both RGB and Euclidian space.
This coherence is obtained by absorbing the form of geodesic distance on the surface of image intensity, then by negative exponential transformation.
The inner coherence $\psi$ can be formalized as:
\vspace{-0.5em}
\begin{equation}
    \begin{aligned}
        \psi(p,q) &= \mathcal{A}_s(p,q) + \lambda_\psi e^{-\mathcal{D}(p,q)},\\
        \mathcal{D}(p,q) &= \min_{P}\int_0^1\|\nabla I\left(P(s)\right)\cdot v(s)\|ds,
    \end{aligned}
    \label{eq:psi}
    \vspace{-0.5em}
\end{equation}
where for pixel $p$ and $q$, $\mathcal{A}_s(p,q)$ is the self-attention and $\mathcal{D}(p,q)$ is the geodesic distance;
$P$ is an arbitrary path from $p$ to $q$ parameterized by $s\in[0,1]$;
$v(s)$ denotes the unit vector $P'(s)/\|P'(s)\|$ that is tangent to the path direction; $I(\cdot)$ is image RGB intensity.

\vspace{0.2cm}
{\noindent \bf Calculating Mask.}
Given objectness and inner coherence, we define an energy function $E$ for each potential mask $\mathcal{M}$:
\vspace{-0.5em}
\begin{equation}
E(\mathcal{M}) = \sum\nolimits_p\phi(p)+\lambda\sum\nolimits_{\mathcal{M}(p)\neq\mathcal{M}(q)}\psi(p,q),
\vspace{-0.5em}
\end{equation}
where $\lambda$ denotes the weight between $\phi$ and $\psi$; $\mathcal{M}(\cdot)\in\{0,1\}$ means the pixel in this mask. The binary mask $\mathcal{M}$ is generated by minimizing $E(\mathcal{M})$, \ie, use Ford-Fulkerson algorithm~\cite{ford1956maximal} to find a minimum cut in the image graph. And after further post-processing and denoising~\cite{barron2016fast, zhang2021datasetgan,li2022bigdatasetgan}, we can obtain the final synthetic mask (see \figref{fig:teaser} Right for some examples).

\vspace{0.2cm}
{\noindent \bf Discussion.}
Compared with other training-free mask generation methods like NCut~\cite{shi2000normalized} and K-means~\cite{lloyd1982},
they only consider pairwise similarly, thus cannot decide fore/background for each partition. 
Compared with DenseCRF~\cite{NIPS2011_beda24c1},
AttentionCut has well-designed objectness and inner coherence terms, which is more suitable for diffusion models and guarantees convergence.
In \tabref{tab:raw_cut}, we have conducted experiments to validate the superiority of AttentionCut.

\subsection{Exploitation Stage: Diffusion Knowledge }
\label{Diffusion Features and Synthetic Data Supervision}

This stage aims to bridge the architectural gap between pre-trained diffusion models and discriminative tasks, \eg, object discovery. 
As shown in \figref{fig:framework} (2), we achieve this in two steps: in \secref{Extracting Diffusion Features}, we treat diffusion models as a universal feature extractor to distill explicit visual knowledge; in \secref{Segment Decoder}, we feed diffusion features into one flexible decoder, and train with ``infinite'' synthetic data.

\subsubsection{Extracting Diffusion Knowledge}
\label{Extracting Diffusion Features}

For diffusion models, they are fed with noise and text to output synthesis images; while for object discovery models, they are fed with images to output pixel-level masks. Such an architectural gap blocks direct feature extraction from diffusion. 
To solve this, given one image, we are required to find the corresponding input noise of diffusion models under some conditioning text, then features can be extracted through diffusion reverse process.
To get input noise, we combine diffusion inversion~\cite{ddim}
with the conditional UNet. To get the conditioning text, we simply classify images by CLIP~\cite{clip}.

\vspace{0.2cm}
{\noindent \bf Diffusion Inversion and Feature Extraction.} 
Given pre-trained diffusion models, we here inverse one image back to its corresponding noise under the conditioning text. 
This diffusion inversion can be seen as a special forward process.

One trivial solution is to use the typical DDPM~\cite{ddpm}. Although it can yield latent variables (\ie, noise) through the forward process, 
these variables are stochastic and cannot reconstruct the image through the reverse process.
So it is not suitable for feature extraction.
Inspired by DDIM~\cite{ddim}, we modify each step by combining it with conditional denoising UNet $\boldsymbol\epsilon_\theta(\boldsymbol{x}_t, t, y)$ in Stable Diffusion, making the forward/reverse non-Markovian to enjoy deterministic. 
Now the forward/reverse process for each step is:
\vspace{-0.5em}
\begin{equation}
\resizebox{0.90\linewidth}{!}{$
\begin{aligned}
    &\boldsymbol{x}_{t+1}=\sqrt{\alpha_{t+1}} \boldsymbol{f}_\theta\left(\boldsymbol{x}_t, t, y\right)+\sqrt{1-\alpha_{t+1}} \boldsymbol{\epsilon}_\theta\left(\boldsymbol{x}_t, t, y\right),
    \\
    &\boldsymbol{x}_{t-1}=\sqrt{\alpha_{t-1}} \boldsymbol{f}_\theta\left(\boldsymbol{x}_t, t,y\right)+\sqrt{1-\alpha_{t-1}} \boldsymbol{\epsilon}_\theta\left(\boldsymbol{x}_t, t,y\right),
    \end{aligned}$
    }
    \label{equ: deterministic reverse}
    \vspace{-0.5em}
\end{equation}
where $\boldsymbol{f}_\theta\left(\boldsymbol{x}_t, t, y\right)=\left({\boldsymbol{x}_t-\sqrt{1-\bar{\alpha}_t} \boldsymbol{\epsilon}_\theta\left(\boldsymbol{x}_t, t, y\right)}\right)\,/\, {\sqrt{\bar{\alpha}_t}}$, $\alpha_t=1-\beta_t$, $\bar{\alpha}_t=\prod_{s=1}^t\left(1-\beta_s\right)$, $\beta_t$ is a variance schedule. $y$ denotes the conditional text, and $t$ means timesteps.

After diffusion inversion, to get the corresponding noise, features $\mathcal{F}^{t,l}$ can be extracted from $\boldsymbol\epsilon_\theta(\boldsymbol{x}_t, t, y)$ during each reverse step $t=T-1, \dots, 0$ and intermediate layer $l = 1, \dots, L$.
To cover long range and multi-level features of multi-scale objects, they are aggregated in all time steps:
\vspace{-0.5em}
\begin{equation}
\mathcal{F}^l=1/T\cdot\sum\nolimits_{t=0}^{T-1} \mathcal{F}^{t,l}.
\vspace{-0.5em}
\end{equation}
In practice, we choose the output of the ``SpatialTransformer'' block in Stable Diffusion, where $L=6$ with resolutions $16 \times 16$, $32 \times 32$, and $64 \times 64$, two of each.

\input{sec/table/main_table}

\vspace{0.2cm}
{\noindent \bf CLIP-classifiable Prior.}
Notice that in \equref{equ: deterministic reverse}, the diffusion inversion should be done under some conditional text $y$.
We choose $y$ to be the CLIP-classified category of the input image, because of the following observations:
(1) humans take pictures by naturally framing an object of interest near the center of the image~\cite{judd2009learning} (center prior); 
(2) most background regions can be easily connected to image boundaries, while difficult for object regions~\cite{wei2012geodesic} (background prior); 
(3) CLIP is pre-trained on a large corpus of web-curated data, and most of which is human-token images with saliency objects~\cite{clip} (source prior). 
It is easy to classify images with the center and background priors, and the source prior enables us to classify using CLIP~\cite{clip}.
We summarize this as \textit{CLIP-classifiable prior}.

In practice, we choose the label set in ImageNet~\cite{imagenet}, and combine semantically similar classes, \eg, poodle and Chihuahua as dogs, etc.
Besides, multiple prompt templates are used, \eg, ``A photo of \{category\}'' to boost performance.

\subsubsection{Segment Decoder}
\label{Segment Decoder}
To enable diffusion compatible with object discovery, we here propose two options for preference. 
One is to attach a flexible decoder to the pre-trained diffusion models, and train using the synthesised data to achieve object discovery. This option costs many parameters and rich training data, bringing superior performance, and we denote it as {\textit{DiffusionSeg}} in  {\tabref{table:main_all}}. 
The other is to extract cross- and self-attention during diffusion inversion, and generate pseudo-masks using AttentionCut in \secref{synthetic mask generation}.
Such an option costs no trainable parameters and data, thus showing faster inference speeds, and we call it {\textit{AttentionCut}} in \tabref{table:main_all}.

\subsection{Discussion}
This paper uses pre-trained diffusion models for unsupervised object discovery.
Comparing with discriminative pre-training~\cite{lost,tokencut,selfmask}, generative pre-training has additional pixel-level understanding, which is more suitable for object discovery. 
Compared with MAE-style~\cite{mae} generative pre-training, which learns reconstruction representations to help object discovery, diffusion models show a clear advantage, \ie, synthesis abundant data, which is valuable to improve performance (see \tabref{tab:train_syn} and \figref{fig:scale}). 
Comparing with GANs in image synthesizing, diffusion models have significant advantages in higher sample quality and diversity, more stability and robustness~\cite{dhariwal2021diffusion}.
Compared to a few early GAN-based works that struggle to synthesise  mask with manual annotations~\cite{zhang2021datasetgan,li2022bigdatasetgan}, diffusion model can obtain mask using AttentionCut, without manually labeling.

%% file: sec/table/main_table.tex
\begin{table*}
  \begin{subtable}{0.57\linewidth}
    \centering
    \resizebox{.98\textwidth}{!}{
    \begin{tabular}{l|ccc|ccc|ccc}
      \toprule
      \multirow{2}{*}{Model}
       &\multicolumn{3}{c|}{DUT-OMRON~\cite{dut_omron}}
       &\multicolumn{3}{c|}{DUTS-TE~\cite{duts}}
       &\multicolumn{3}{c}{ECSSD~\cite{ecssd}}\\   \cmidrule{2-10}
       &Acc$\uparrow$ &IoU$\uparrow$ &max$F_\beta$$\uparrow$
       &Acc$\uparrow$ &IoU$\uparrow$ &max$F_\beta$$\uparrow$
       &Acc$\uparrow$ &IoU$\uparrow$ &max$F_\beta$$\uparrow$\\
      \midrule
      HS~\cite{yan2013cvpr}
      &.843&.433&.561
      &.826&.369&.504
      &.847&.508&.673\\
      wCtr~\cite{zhu2014saliency}
      &.838&.416&.541
      &.835&.392&.522
      &.862&.517&.684\\
      WSC~\cite{li2015cvpr}
      &.865&.387&.523
      &.862&.384&.528
      &.852&.498&.683\\
      DeepUSPS~\cite{Nguyen2019usps}
      &.779&.305&.414
      &.773&.305&.425
      &.795&.440&.584\\
      BigBiGAN~\cite{voynov2021object}
      &.856&.453&.549
      &.878&.498&.608
      &.899&.672&.782\\
      E-BigBiGAN~\cite{voynov2021object}
      &.860&.464&.563
      &.882&.511&.624
      &.906&.684&.797\\
      Melas-Kyriazi et al.~\cite{melas2021finding}
      & .883 & .509 & -
      & .893 & .528 & - 
      &.915 & .713 & -\\
      LOST~\cite{lost}
      &.797&.410&.473
      &.871&.518&.611
      &.895&.654&.758\\
      Deep Spectral~\cite{MelasKyriazi2022DeepSM}
      &- &.567&-
      &- &.514&-
      &- &.733&-\\
      TokenCut~\cite{tokencut}
      &.880&.533&.600
      &.903&.576&.672
      &.918&.712&.803\\
      FreeSOLO~\cite{Wang2022FreeSOLOLT}
      &.909&.560&.684
      &.924&.613&{.750}
      &.917&.703&.858\\
      SelfMask (pseudo)~\cite{selfmask}
      &.811&.403&-
      &.845&.466&-
      &.893&.646&-\\
      SelfMask~\cite{selfmask}
      &.901&.582&.680
      &.923&.626&.750
      &.944&.781&.889\\
      FOUND-single~\cite{simeoni2022unsupervised}
      &.920&.586&.683
      &.993&.637&.733
      &.912&.793&.946\\
      FOUND-multi~\cite{simeoni2022unsupervised}
      &.912&.578&.663
      &.938&.645&.715
      &.949&.807&.955\\
      \midrule
      LOST~\cite{lost} +BS
      &.818&.489&.578
      &.887&.572&.697
      &.916&.723&.837\\
      TokenCut~\cite{tokencut} +BS
      &.897&.618&.697
      &.914&.624&.755
      &.934&.772&.874\\
      SelfMask~\cite{selfmask} +BS
      &.919&.655&.771
      &.933&.660&.819
      &.955&.818&.911\\
      FOUND-single~\cite{simeoni2022unsupervised} +BS
      &.921&.608&.706
      &.941&.654&.733
      &.912&.793&.946\\
      FOUND-multi~\cite{simeoni2022unsupervised} +BS
      &.922&.613&.708
      &.942&.663&.763
      &.951&.813&.935\\
      \midrule
      \textbf{AttentionCut}
      &.905&.536&-
      &.914&.608&-
      &.924&.710&-\\
      \textbf{\methodName}
      &\textbf{.948} &\textbf{.661} &\textbf{.772}
      &\textbf{.959} &\textbf{.704} &\textbf{.829}
      &\textbf{.964} &\textbf{.831} &\textbf{.955}\\
      \bottomrule
      \end{tabular}}
    \caption{\textbf{Comparisons for unsupervised saliency segmentation} on three standard benchmarks DUT-OMRON~\cite{dut_omron}, DUTS~\cite{duts} and ECSSD~\cite{ecssd}. +BS means Bilateral Solver~\cite{barron2016fast} for post-processing. The max$F_\beta$ on SelfMask has been re-evaluated for fair comparisons, as we found SelfMask computed max$F_\beta$ with various optimal thresholds, while other methods only use one unified threshold. 
    }
    \label{tab:main_seg}
  \end{subtable}
    \begin{subtable}{0.42\linewidth}
      \centering
      \resizebox{.98\textwidth}{!}{
      \begin{tabular}{l|ccc}
        \toprule Method &  VOC07~\cite{pascal-voc-2007} & VOC12 ~\cite{pascal-voc-2012}& COCO20K~\cite{Lin2014cocodataset,Vo20rOSD}  \\
        \midrule
        Selective Search~\cite{uijlings2013selective, lost} & 18.8 & 20.9 & 16.0 \\
        EdgeBoxes~\cite{zitnick2014edge, lost} & 31.1 & 31.6 & 28.8 \\
        Kim et al.~\cite{kim2009pagerank_uod, lost}&  43.9 & 46.4& 35.1 \\
        Zhange et al.~\cite{zhang2020object, lost}&  46.2 & 50.5 & 34.8 \\
        DDT+~\cite{Wei2019ddtplus, lost}&  50.2 & 53.1 & 38.2 \\
        rOSD~\cite{Vo20rOSD, lost} &  54.5 & 55.3 & 48.5 \\
        LOD~\cite{vo2021largescale, lost}&53.6 & 55.1 & 48.5 \\
        DINO-seg~\cite{lost}&   45.8 & 46.2 & 42.1 \\
        FreeSOLO~\cite{Wang2022FreeSOLOLT} &  56.1 & 56.7 & 52.8 \\
        LOST~\cite{lost}& 61.9 & 64.0 & 50.7 \\
        Deep Spectral \cite{MelasKyriazi2022DeepSM} & 62.7 & 66.4 & 52.2 \\
        TokenCut \cite{tokencut}& 68.8 & 72.1 &  58.8 \\ 
        \midrule
        \textbf{AttentionCut} & 67.5 & 70.2 & 54.9 \\
        \textbf{\methodName} & \textbf{75.2} & \textbf{78.3} & \textbf{63.1} \\
        \bottomrule
      \end{tabular}}
      \caption{\textbf{Single object localization.} We extract the tight bounding box for saliency mask as our box prediction.}
      \label{tab:main_det}
      \centering
      \resizebox{.65\textwidth}{!}{
              \begin{tabular}{l|ccc}
              \toprule
              Model
              &Acc &IoU &max$F_\beta$ \\\midrule
              PertGAN~\cite{bielski2019emergence}
              &- &.380 &-\\
              ReDO~\cite{chen2019unsupervised}
              &.845 &.426 &-\\
              OneGAN~\cite{benny2020onegan}
              &- &.555 &-\\
              Melas-Kyriazi~\cite{melas2021finding}
              &.921 &.664 &.783\\
              BigBiGAN~\cite{voynov2021object}
              &.930 &.683 &.794\\
              E-BigBiGAN~\cite{voynov2021object}
              &.940 &.710 &.834\\
              \midrule
              \textbf{AttentionCut}
              &.946 &.695 &.838\\
              \textbf{\methodName}
              &\textbf{.963} &\textbf{.726} &\textbf{.852}\\
              \bottomrule
            \end{tabular}
        }
      \caption{\textbf{Compare with GAN-based methods on CUB~\cite{cub}}.}
      \label{table:bird}
    \end{subtable}
    \vspace{-0.2cm}
  \caption{\textbf {Comparison with state-of-the-art methods on object discovery.} 
  Our \methodName outperforms previous state-of-the-art approaches across all benchmarks.}
  \label{table:main_all}
\end{table*}

%% file: sec/4-experiments.tex
\section{Experiments}
\subsection{Experimental Setup}
{\noindent \bf Datasets \& Evaluations.}
For unsupervised saliency segmentation, we evaluate on three standard benchmarks: ECSSD~\cite{ecssd}, DUTS~\cite{duts} and DUT-OMRON~\cite{dut_omron}. We also use CUB~\cite{cub} to compare with some generative-based segmentation models~\cite{chen2019unsupervised, voynov2021object, melas2021finding}.
For metrics, we report pixel-wise accuracy (Acc), intersection-over-union (IoU), and max$F_\beta$ for $\beta^2$ to $0.3$ following conventions~\cite{voynov2021object, melas2021finding, tokencut, selfmask}.

For unsupervised single object localization, we evaluate on VOC07~\cite{pascal-voc-2007}, VOC12~\cite{pascal-voc-2012} and COCO20K~\cite{Lin2014cocodataset,Vo20rOSD}. We evaluate using correct localization (CorLoc)~\cite{tokencut}, \ie, the percentage of images, where the IoU $> 0.5$ of a predicted single bounding box with at least one of the ground truth.

\input{sec/table/data_statistics}

\vspace{0.2cm}
{\noindent \bf Implementation Details.}
We adopt the publicly released \texttt{sd-v1-4.ckpt} of Stable Diffusion\footnote{https://huggingface.co/CompVis/stable-diffusion-v-1-4-original} for image generation, and it remains frozen throughout. 
We set image resolution $512\times 512$, timesteps $T=40$, channel num $C=4$, sample frequency $f=8$ and ddim eta is $0.0$. 
For AttentionCut, we set $\lambda_{\phi}=0.16$, $\lambda_{\psi}=2.5$ and $\lambda=0.1$.
Our synthetic dataset contains about $50,000$ image-mask pairs.
We use a three-layer FCN as segment decoder, and set $\text{lr}=0.001$ on Adam~\cite{kingma2014adam} for optimization. The batch size is set to $10$.

\subsection{Comparison with the State-of-the-art}
{\noindent \bf Unsupervised Saliency Segmentation.} 
\tabref{tab:main_seg} compares for unsupervised object segmentation.
\methodName
reached a new SOTA, and largely improves AttentionCut by 12.5\%, 9.6\%, 12.1\% in IoU after training on our synthetic dataset, which proves the value of synthesis data.

Besides, we also compared with some GAN-based unsupervised object segmentation methods on CUB benchmark, as shown in \tabref{table:bird}.
Our \methodName, utilizing diffusion model, can largely outperform all GAN-based method even without the need of training on the synthetic data.

\vspace{0.1cm}
{\noindent \bf Unsupervised Single Object Localization.}
Given the predicted segmentation mask from our model, we convert it to a bounding box by first connecting components, then choosing the tight outline of the largest components from the top, bottom, left and right sides. As shown in \tabref{tab:main_det}, our method reaches new state-of-the-art on all three benchmarks.

\subsection{Synthesized Data Analysis}
This section provides a thorough analysis of our synthesized dataset. The results show that with sufficient data scale, our dataset is a reliable simulation of the real world.

\subsubsection{Data Statistics}
We compare our synthetic dataset to a real dataset DUTS-TR~\cite{duts}. For most key properties, statistics show that our synthetic dataset has a similar distribution to DUTS-TR.

\vspace{0.2cm}
{\noindent \bf Color Contrast.}
As shown in~\figref{fig:color}, our dataset has almost the same distribution of color contrast as DUTS-TR, which can make the model easy to transfer to the real world.

\vspace{0.2cm}
{\noindent \bf Object Size.}
Defining object size as the ratio of foreground pixels to full image pixels, \figref{fig:size} shows the comparison.
Compared to DUTS-TR, 
our synthetic data has a broader scale of salient objects (object sizes ranging from $0.1$ to $0.5$), which is suitable for training the object discovery task.

\vspace{0.2cm}
{\noindent \bf Center Bias.}
\figref{fig:center_bias} draws the scatter plot for each object using bounding box centers. 
In comparison with object-centric DUTS-TR, a more diverse center distribution contains more hard samples, and can improve generalizability of the model.

\vspace{0.2cm}
{\noindent \bf Geometry Statistics.}
\tabref{tab:shape} shows shape complexity (SC), polygon length (PL) and shape diversity (SD) of our dataset, which are close with real DUTS-TR.
Following~\cite{li2022bigdatasetgan}, we convert masks into polygons and define SC as vertice number, PL as perimeter. SD is defined as averaging pairwise Chamfer distance between two polygons.

\subsubsection{Training Performance}
We train two typical segmentation pipeline on different scale of synthetic dataset as well as real ones. Synthetic data is a replacement for real data with sufficient scale.

\vspace{0.2cm}
{\noindent \bf Compared with Real Dataset.}
Ideally, a well-established dataset should be capable of training on arbitrary architectures. To reveal such ability, we compare performances of training on our synthetic dataset and DUTS-TR. Specifically, we deal with saliency segmentation in an end-to-end manner. We select two widely used segmentation architectures, UNet~\cite{ronneberger2015u} and DeepLabV3~\cite{Chen2017RethinkingAC} as representatives. \textbf{Note that}, training on real data makes use of both image and \textit{ground truth} annotations, which can be seen as \textit{fully supervised} in this scenario.

\input{sec/table/train_on_syn}
\tabref{tab:train_syn} shows experimental results. The gap between synthetic and real data can be observed on the same data scale. However, performance could be boosted by adding more synthetic data. With an increase of $10\times$ in scale, the model is trained to be comparable with that on real data. Considering the infinite generating capability of diffusion models, we come to the conclusion that our synthetic data is a viable alternative to real ones.

\vspace{0.2cm}
{\noindent \bf Data Scale.}
~\figref{fig:scale} answers for ``how much synthetic data is enough for training?'' We increase the data scale from $1k$ to $100k$ and report IoU on DUTS-TE, and find there's a decrease marginal effect as scale increases. We keep the scale to $50k$, for its good trade-off between synthesizing cost and performance.

\input{sec/table/scale}

\subsection{Diffusion Features Analysis}
\input{sec/table/diffusion_feature}

To show the superiority of pre-trained diffusion features, we compare them with some discriminative pre-training methods (DINO~\cite{dino}, MoCo~\cite{moco}, CLIP~\cite{clip}). 
During training, we freeze all pre-trained models as feature extractor, attaching a same segment decoder for mask prediction.

\vspace{0.2cm}
{\noindent \bf Diffusion Pre-training \textit{vs.} Discriminative Pre-training.}
\tabref{tab:diffusion_feature} shows the privilege of diffusion features, compared with discriminative ones. Noticing that pixel-wise reconstruction is the most informative pre-training task among the three, it's not surprising to have such results. 
Although both diffusion and CLIP are pre-trained using text-image pairs,
discriminative pre-training like image-caption alignment focuses mainly on global features and loses detail.

\subsection{Ablation Study}

\vspace{0.2cm}
{\noindent \bf Mask Generation Methods.}
Besides our AttentionCut, \tabref{tab:raw_cut} also compares with other training-free segmentation methods.
They are usually used in RGB space.
Here, we apply them to diffusion features with minor modifications.
Overall, AttentionCut far outperforms these methods.

\input{sec/table/raw_cut}
\input{sec/table/attentioncut}

\vspace{0.2cm}
{\noindent \bf AttentionCut Components.}
\tabref{tab:attentioncut} ablates the AttentionCut formulation to show their effectiveness. $\mathcal{A}_c$ alone (\ding{172}) provides a reasonable mask prediction and can be enhanced by $r(p)$ (\ding{173},\,\ding{174}).
Both semantic and spatial coherence improve results (\ding{175},\,\ding{176}).
The benefits of two coherences are addable, and combining them all performs best (\ding{177}).

\vspace{0.2cm}
{\noindent \bf Effectiveness of CLIP-classifiable Prior.}
\input{sec/table/classification_prior}
~\tabref{tab:class_prior} ablates the CLIP-classifiable prior. Under \textit{w/o prior} setting, the category label is replaced by an empty string.
It shows AttentionCut heavily relies on this prior, but
DiffusionSeg is robust.
As AttentionCut is based on attention maps, it is sensitive to the given label. 
However, in DiffusionSeg, the network is trained, potentially having the ability to understand diffusion features \textit{without} CLIP.

%% file: sec/table/data_statistics.tex
\begin{table*}
    \begin{minipage}[t]{0.19\linewidth}
        \vspace{0pt}
        \centering
        \includegraphics[width=0.98\linewidth]{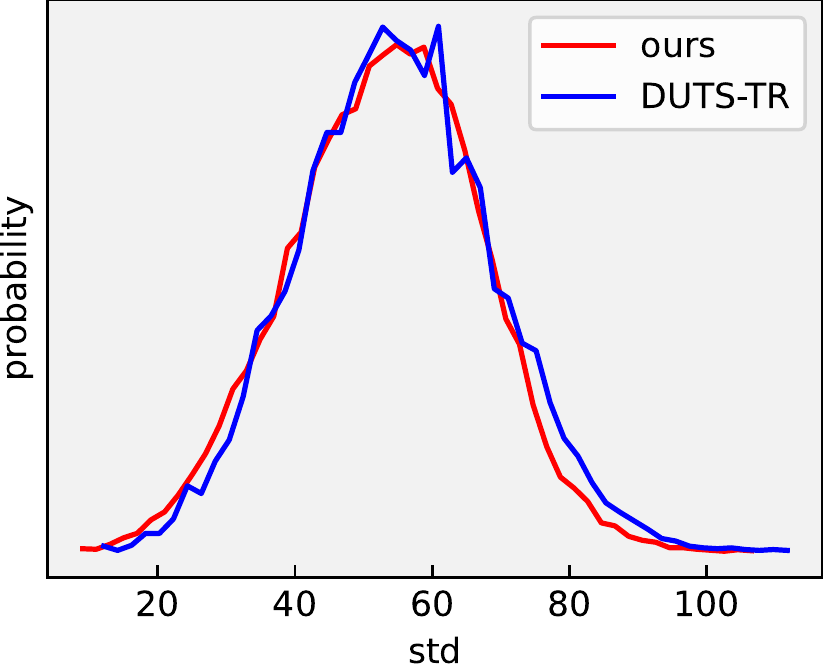}
        \captionsetup{font={small}}
        \figcaption{\textbf{Color contrast.} Our synthetic data shows a similar distribution of {color contrast} with real-world dataset DUTS-TR.}
        \label{fig:color}
    \end{minipage}\hspace{5pt}
    \begin{minipage}[t]{0.19\linewidth}
        \vspace{0pt}
        \centering
        \includegraphics[width=0.98\linewidth]{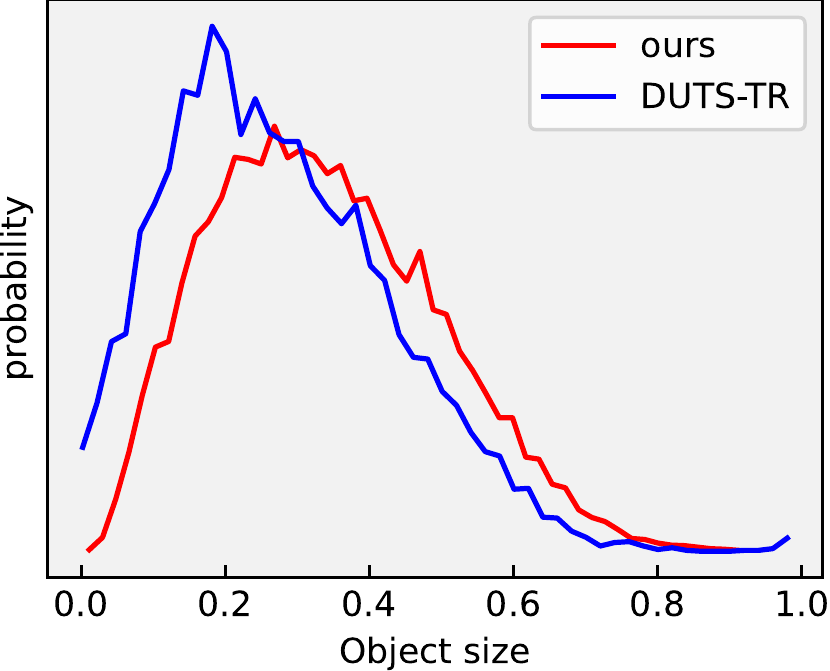}
        \captionsetup{font={small}}
        \figcaption{\textbf{Object size.} 
        Our synthetic data has a broader scale of salient objects (object sizes ranging from $0.1$ to $0.5$).
        }
        \label{fig:size}
    \end{minipage}\hspace{5pt}
    \begin{minipage}[t]{0.37\linewidth}
            \vspace{0pt}
            \includegraphics[width=.48\linewidth]{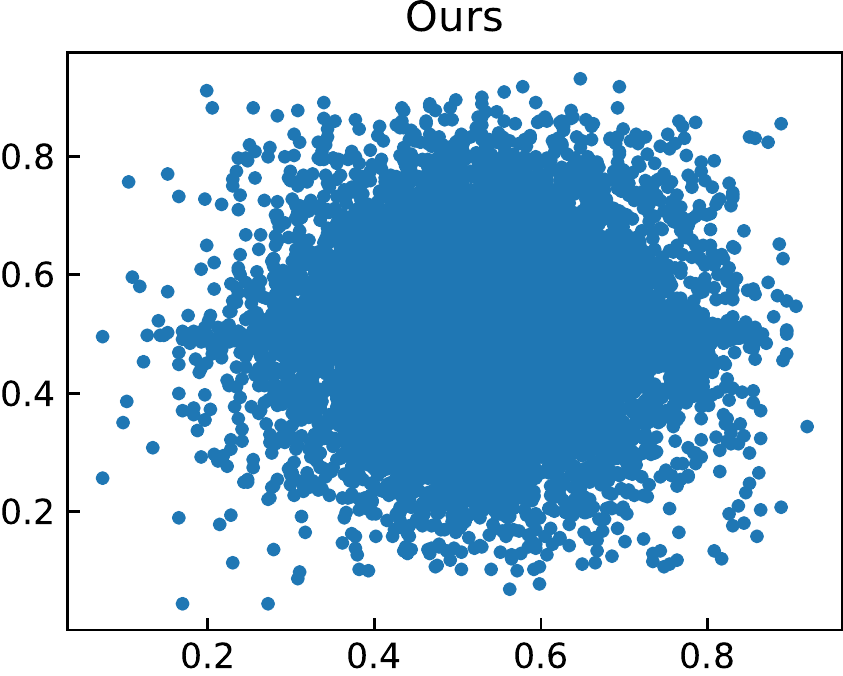}
            \includegraphics[width=.48\linewidth]{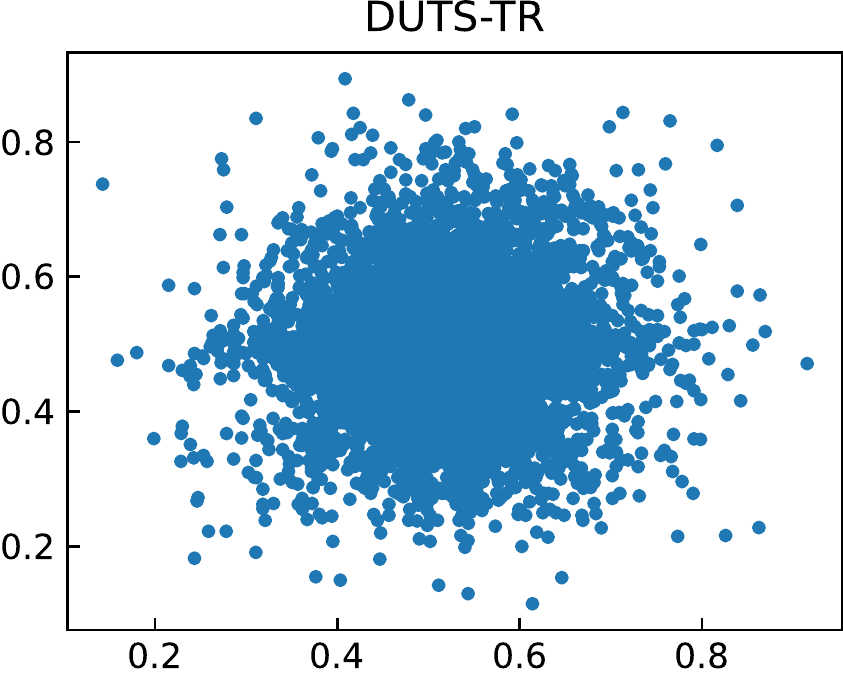}
            \figcaption{\textbf{Center bias scatter plot for our synthetic dataset (left) and DUTS-TR (right).} DUTS-TR is object-centric, while our dataset is more diverse in center distribution and contains more hard samples.}
            \label{fig:center_bias}
    \end{minipage}\hspace{5pt}
    \begin{minipage}[t]{0.19\linewidth}
        \vspace{0pt}
        \resizebox{.98\textwidth}{!}{
          \begin{tabular}{c|cc}
            \toprule
            &DUTS-TR&Ours\\\midrule
            SC&27.1&29.9\\\midrule
            PL&2.96&2.78\\\midrule
            SD&2.31&1.91\\
            \bottomrule
          \end{tabular}
        }
        \tabcaption{\textbf{Geometry statistics,} in terms of shape complexity (SC), polygon length (PL) and shape diversity (SD). 
        }
        \label{tab:shape}
      \end{minipage}
\end{table*}

%% file: sec/table/train_on_syn.tex
\begin{table}[!ht]
  \small
    \centering
    \resizebox{.48\textwidth}{!}{
          \begin{tabular}{c|c|cc|cc|cc}
          \toprule
          \multirow{2}{*}{Model}
          &\multirow{2}{*}{Dataset}
           &\multicolumn{2}{c|}{DUT-OMRON}
           &\multicolumn{2}{c|}{DUTS-TE}
           &\multicolumn{2}{c}{ECSSD}\\  \cmidrule{3-8}
           &&Acc &IoU
           &Acc &IoU
           &Acc &IoU\\
          \midrule
          \multirow{3}*{UNet~\cite{ronneberger2015u}}
          &{Real $1k$}
          &.853&.471
          &\textbf{.875}&.489
          &\textbf{.912}&.678\\
          &{Syn $1k$}
          &.841&.419
          &.846&.401
          &.853&.615\\
          &{Syn $10k$}
          &\textbf{.864}&\textbf{.475}
          &.872&\textbf{.493}
          &.908&\textbf{.681}\\\midrule
          \multirow{3}*{DeepLabV3~\cite{Chen2017RethinkingAC}}
          &{Real $1k$}
          &.910&.565
          &.900&.512
          &.899&\textbf{.670}\\
          &{Syn $1k$}
          &.878&.479
          &.866&.454
          &.862&.612\\
          &{Syn $10k$}
          &\textbf{.916}&\textbf{.573}
          &\textbf{.906}&\textbf{.521}
          &\textbf{.904}&.669\\
          \bottomrule
          \end{tabular}
    }
  \caption{\textbf{Performance of segmentation models trained on synthetic and real dataset.} Real 1k refers to 1000 random selected images from DUTS-TR. It shows synthetic data can replace real data with sufficient samples provided.}
  \vspace{-1.1em}
  \label{tab:train_syn}
\end{table}

%% file: sec/table/scale.tex
\begin{table}
    \begin{minipage}[m]{0.5\linewidth}
      \centering
      \includegraphics[width=.98\linewidth]{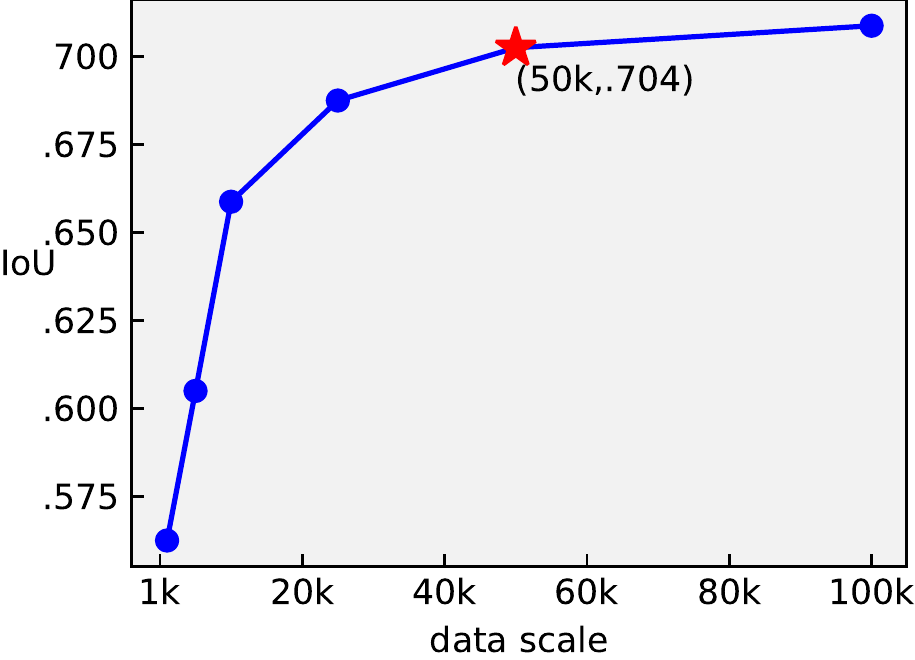}
    \end{minipage}
    \begin{minipage}[m]{0.49\linewidth}
      \figcaption{\textbf{Results of increasing the data scale from $1k$ to $100k$ on DUTS-TE.} 
      The data scale of $50k$ is finally used, since it offers a goodish trade-off between synthetic cost and performance at the same time.}
      \label{fig:scale}
  \end{minipage}
      \vspace{-1.2em}
  \end{table}

%% file: sec/table/diffusion_feature.tex
\begin{table}[!ht]
    \small
      \centering
      \resizebox{.48\textwidth}{!}{
            \begin{tabular}{c|c|cc|cc|cc}
            \toprule
            \multirow{2}{*}{Model}
            &\multirow{2}{*}{Dataset}
             &\multicolumn{2}{c|}{DUT-OMRON}
             &\multicolumn{2}{c|}{DUTS-TE}
             &\multicolumn{2}{c}{ECSSD}\\  \cmidrule{3-8}
             &&Acc &IoU
             &Acc &IoU
             &Acc &IoU\\
            \midrule
            DINO~\cite{dino}&\multirow{4}*{{Syn $5k$}}
            &.897&.528
            &.883&.502
            &.869&.701\\
            MoCo~\cite{moco}
            &&.882&.534
            &.912&.596
            &.915&.699\\
            CLIP~\cite{clip}
            &&\textbf{.897}&.523
            &\textbf{.924}&.563
            &.920&.728\\
            \textbf{Ours}
            &&.895&\textbf{.532}
            &.916&\textbf{.594}
            &\textbf{.923}&\textbf{.730}\\\midrule
            DINO~\cite{dino}&\multirow{4}*{{Syn $10k$}}
            &.930&.589
            &.919&.576
            &.907&.750\\
            MoCo~\cite{moco}
            &&.925&.586
            &.915&.623
            &.930&.766\\
            CLIP~\cite{clip}
            &&.931&.592
            &.913&.628
            &.933&.771\\
            \textbf{Ours}
            &&\textbf{.932}&\textbf{.598}
            &\textbf{.933}&\textbf{.637}
            &\textbf{.952}&\textbf{.790}\\
            \bottomrule
            \end{tabular}
      }
    \caption{\textbf{Comparing diffusion models (Ours) with other pre-trained models.} Diffusion features show privilege.}
    \label{tab:diffusion_feature}
    \vspace{-1.2em}
\end{table}

%% file: sec/table/raw_cut.tex
\begin{table}[!ht]
    \small
      \centering
      \resizebox{.98\linewidth}{!}{
      \begin{tabular}{c|cc|cc|cc}
        \toprule
        \multirow{2}{*}{Method}
        &\multicolumn{2}{c|}{DUT-OMRON}
        &\multicolumn{2}{c|}{DUTS-TE}
        &\multicolumn{2}{c}{ECSSD}\\
        \cmidrule{2-7}
        &Acc &IoU&Acc &IoU&Acc &IoU\\\midrule
        K-means clustering~\cite{lloyd1982} &.802 &.413 &.834 &.462 &.885 &.628\\
        DenseCRF~\cite{NIPS2011_beda24c1} &.872 &.497 &.883 &.522 &.902&.692\\
        NCut~\cite{shi2000normalized} &.860 &.503 &.872 &.528 &.899 &.690\\
        \textbf{AttentionCut}&\textbf{.905} &\textbf{.536} &\textbf{.914} &\textbf{.608} &\textbf{.924} &\textbf{.710}\\
        \bottomrule
      \end{tabular}
    }
    \vspace{-0.5em}
    \caption{\textbf{Comparisons of training-free mask generation methods.} All are conducted in diffusion feature space.}
    \label{tab:raw_cut}
\end{table}

%% file: sec/table/attentioncut.tex
\begin{table}[!ht]
    \small
      \centering
      \resizebox{.49\textwidth}{!}{
            \begin{tabular}{c|cc|cc|
            cc|cc|cc}
            \toprule
            \multirow{2}{*}{Model}
            &\multicolumn{2}{c|}{$\phi(p)$}&\multicolumn{2}{c|}{$\psi(p,q)$}
             &\multicolumn{2}{c|}{DUT-OMRON}
             &\multicolumn{2}{c|}{DUTS-TE}
             &\multicolumn{2}{c}{ECSSD}\\\cmidrule{2-11}
            &$\mathcal{A}_c$&$r(p)$&$\mathcal{A}_s$&$\mathcal{D}$
             &Acc &IoU
             &Acc &IoU
             &Acc &IoU\\
            \midrule
            \ding{172} &\checkmark &- &- &-
            &.881 &.480
            &.902 &.535
            &.903 &.637\\
            \ding{173} &- &\checkmark &- &-
            &.885 &.486
            &.896 &.533
            &.915 &.691\\
            \ding{174} &\checkmark &\checkmark &- &-
            &.892 &.502
            &.910 &.561
            &.905 &.643\\
            \ding{175} &\checkmark &\checkmark &- &\checkmark
            &.896 &.512
            &.910 &.582
            &.919 &.699\\
            \ding{176} &\checkmark &\checkmark &\checkmark &-
            &.894 &.508
            &.909 &.579
            &.912 &.657\\
            \ding{177} &\checkmark &\checkmark &\checkmark &\checkmark
            &\textbf{.905} &\textbf{.536}
            &\textbf{.914} &\textbf{.608}
            &\textbf{.924} &\textbf{.710}\\\bottomrule
            \end{tabular}
      }
      \vspace{-0.5em}
    \caption{\textbf{Ablation on the components of AttentionCut.} $\mathcal{D}$ means spatial coherence in Eq.~\ref{eq:psi}.}
    \label{tab:attentioncut}
    \vspace{-.5em}
\end{table}

%% file: sec/table/classification_prior.tex
\begin{table}[!ht]
    \small
      \centering
      \resizebox{.49\textwidth}{!}{
            \begin{tabular}{c| 
            cc|cc|cc}
            \toprule
            \multirow{2}{*}{Method}
             &\multicolumn{2}{c|}{DUT-OMRON}
             &\multicolumn{2}{c|}{DUTS-TE}
             &\multicolumn{2}{c}{ECSSD}\\   \cmidrule{2-7}
             &Acc &IoU
             &Acc &IoU
             &Acc &IoU\\
            \midrule
            AttentionCut (\it{w/} prior)
            &\textbf{.905} &\textbf{.536}
            &\textbf{.914} &\textbf{.608}
            &\textbf{.924} &\textbf{.710}\\
            AttentionCut (\it{w/o} prior)
            &.831 &.392
            &.816 &.329
            &.851 &.466\\
            \methodName (\it{w/} prior)
            &\textbf{.948} &\textbf{.661}
            &\textbf{.959} &\textbf{.704}
            &\textbf{.964} &\textbf{.831}\\
            \methodName (\it{w/o} prior)
            &.929 &.628
            &.943 &.672
            &.952 &.801\\
            \bottomrule
            \end{tabular}
      }
    \caption{\textbf{Ablation on CLIP-classification prior.} \textit{w/o prior} means using empty string in place of category label.}
    \label{tab:class_prior}
\end{table}

%% file: sec/5-conclusion.tex
\section{Conclusion}

Diffusion model has shown remarkable success on generative tasks. 
In this paper, we propose \methodName to further explore its ability on discriminative tasks.
We build a synthetic dataset using AttentionCut to generate image-mask pairs, and use diffusion inversion to exploit diffusion features for training a segment decoder. Our \methodName shows privilege and achieves new SOTA on all benchmarks.
We expect our work to make a positive step towards unifying generative and discriminative tasks in one model.

%% file: sec/6-appendix.tex
\onecolumn
\section{Appendix}

In this supplementary material, we start by giving details about evaluation metrics in \secref{Evaluation Metrics}, and datesets in \secref{Datesets Details}.
In \secref{Visualization}, some qualitative visualizations about four benchmarks and our synthetic dataset are displayed.

\subsection{Evaluation Metrics}
\label{Evaluation Metrics}
\subsubsection{Saliency Segmentation Metrics}
Here we define three metrics used for evaluating saliency segmentation performance:
\begin{itemize}
    \item \textbf{Accuracy (Acc)} measures pixel-wise accuracy using ground-truth masks $\mathcal{G}\in\{0,1\}^{H \times W}$ and binary predictions $\mathcal{M}\in\{0,1\}^{H \times W}$.
    \begin{equation}
        \text{Acc}=\frac{1}{HW}\sum_{i=1}^H\sum_{j=1}^W\mathbb{I}({\mathcal{G}_{ij}=\mathcal{M}_{ij}}),
    \end{equation}
    where $\mathbb{I(\cdot)}$ is the indicator function.
    \item \textbf{Intersection-over-union (IoU)} is the overlapped size divided by the total size of the foreground regions from $\mathcal{G}$ and $\mathcal{M}$.
    \begin{equation}
        \text{IoU}=\frac{|\mathcal{M} \cap \mathcal{G}|}{|\mathcal{M} \cup \mathcal{G}|}.
    \end{equation}
    \item \textbf{maximal-$F_\beta$ (max$F_\beta$)} is the maximum score of $F_\beta$ among masks binarised using different thresholds. Given binarized mask $\mathcal{M}$ and ground-truth $\mathcal{G}$, $F_\beta$ is defined as:
    \begin{equation}
        F_\beta = \frac{(1+\beta^2)\text{Precision}\times\text{Recall}}{\beta^2\text{Precision}+\text{Recall}},
    \end{equation}
    where $\text{Precision}=\frac{tp}{tp+fp}$ and $\text{Recall}=\frac{tp}{tp+fn}$. $tp,fp,fn$ represent true-positive, false-positive and false-negative respectively. $\beta^2$ denotes weight. We set $\beta^2=0.3$ in our experiments following~\cite{voynov2021object, melas2021finding, tokencut, selfmask}.
\end{itemize}

\subsubsection{Single Object Localization Metrics}
We report performance using \textit{CorLoc} metric following~\cite{tokencut}. CorLoc considers a predicted bounding box to be correct if the intersection over union (IoU) score between this box and one of the ground-truth bounding boxes is greater than $0.5$.

\subsubsection{Geometry Metrics}
In \tabref{tab:shape} we use three metrics to measure the dataset's geometry statistics. Here we provide the implementation details of the three metrics: shape complexity (SC), polygon length (PL) and shape diversity (SD).

Following ~\cite{li2022bigdatasetgan}, we use OpenCV's \texttt{findContours} function with \texttt{RETR\_EXTERNAL} and \texttt{CHAIN\_APPROX\_SIMPLE} flag to extract a simplified polygon for each mask. Then we normalize the polygon by $p_i=(p_i-p_{min})/(p_{max}-p_{min})$. This operation normalizes the polygon to a unit square in both horizontal and vertical directions. $p_{min}$ and $p_{max}$ are the minimum and maximum coordinates among the set of points. We further apply Douglas-Peucker algorithm~\cite{douglas1973algorithms} with a threshold of $0.01$ to simplify the polygon. After that, we define:
\begin{itemize}
    \item \textbf{Shape Complexity (SC)} is the number of points in the normalized and simplified polygon.
    \item \textbf{Polygon Length (PL)} is defined as the total length of the polygon.
    \item \textbf{Shape Diversity (SD)} is the average mean of pair-wise Chamfer distance~\cite{fan2017point} over all dataset. Chamfer distance is:
\begin{equation}
    d_{\text{CD}}(S_1,S_2)=\sum_{p\in S_1}\min_{q\in S_2}\|p-q\|_2^2+\sum_{q\in S_2}\min_{p\in S_1}\|p-q\|_2^2,
\end{equation}
where $S_1$ and $S_2$ are two sets of points corresponding to different polygons.
\end{itemize} 

\subsection{Datesets Details}
\label{Datesets Details}
Here we present details of all benchmarks used in our experiments:
\begin{itemize}
    \item \textbf{ECSSD} (Extended Complex Scene Saliency Dataset)~\cite{ecssd} consists of 1,000 real-world images of complex scenes. 
    \item \textbf{DUT-OMRON}~\cite{dut_omron} contains 5,168 high quality images with very challenging scenarios. 
    \item \textbf{DUTS}~\cite{duts} contains 10,553 training images (DUTS-TR) collected from the ImageNet~\cite{imagenet} DET training/val sets, and 5,019 test images (DUTS-TE) collected from the ImageNet DET test set and the SUN~\cite{sun} dataset. Following previous works~\cite{shen2022learning, tokencut, selfmask}, the performance is reported only on DUTS-TE.
    \item \textbf{CUB} (Caltech-UCSD Birds-200-2011)~\cite{cub} contains 11,788 images and segmentation masks of 200 subcategories belonging to birds. We follow~\cite{chen2019unsupervised, voynov2021object, melas2021finding}  but only use the 1,000 images for the test subset from splits provided by~\cite{chen2019unsupervised}.
    \item \textbf{VOC07}~\cite{pascal-voc-2007} and \textbf{VOC12}~\cite{pascal-voc-2012} correspond to the training and validation set of PASCAL VOC07 and PASCAL VOC12. VOC07 and VOC12 contains 5,011 and 11,540 images respectively which belong to 20 categories.
    \item \textbf{COCO20K} contains 19,817 randomly chosen images from the COCO2014 dataset~\cite{Lin2014cocodataset}. It is used as a benchmark in~\cite{Vo20rOSD} for a large scale evaluation.
\end{itemize}

\clearpage
\subsection{Visualization}
\label{Visualization}
\begin{figure*}[!ht]
    \centering
    \includegraphics[width=0.33\textwidth]{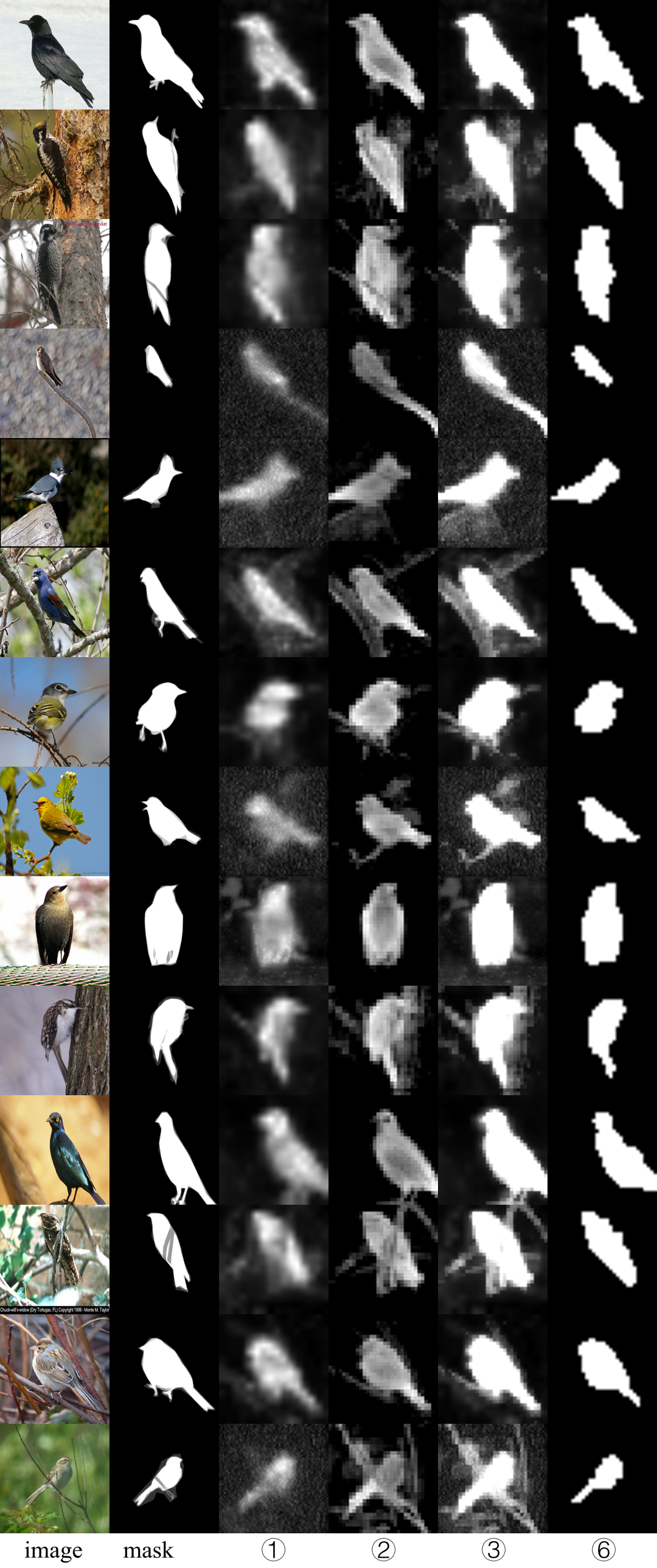}
    \includegraphics[width=0.33\textwidth]{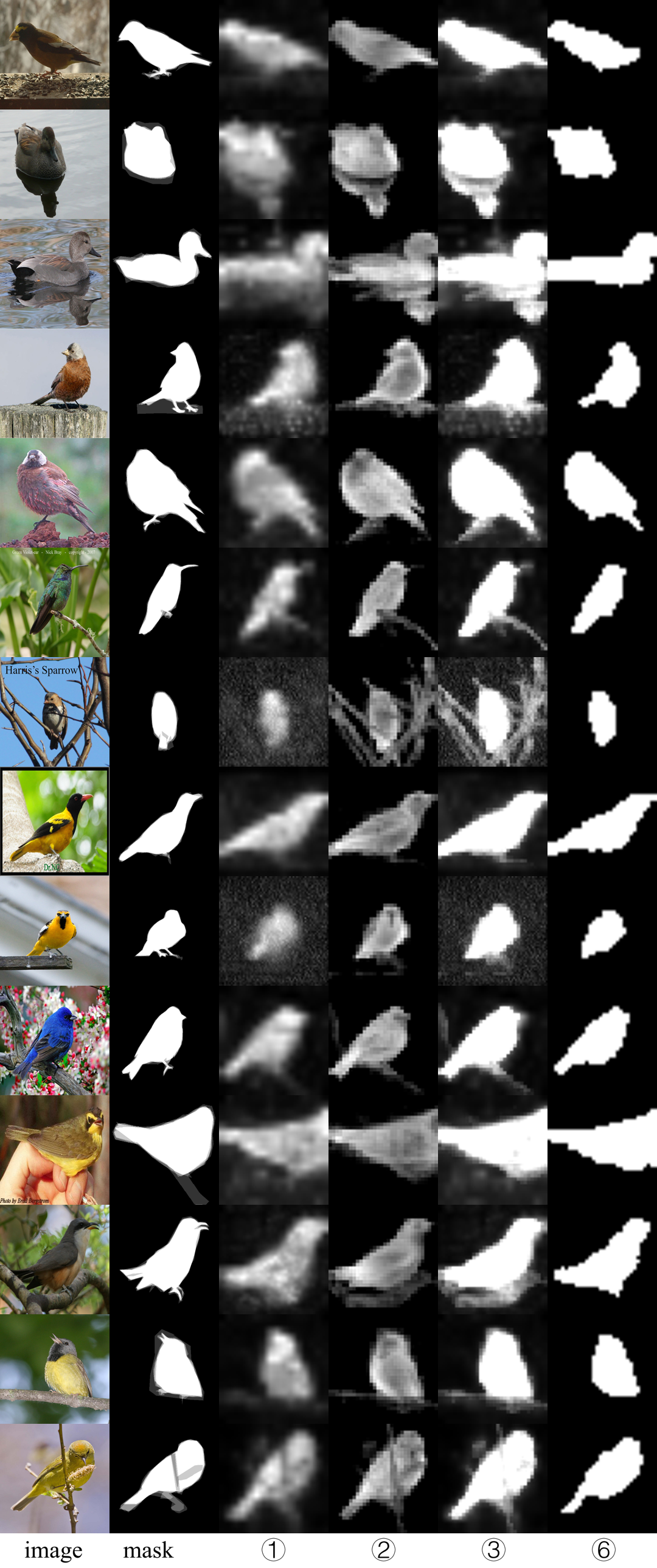}
    \includegraphics[width=0.33\textwidth]{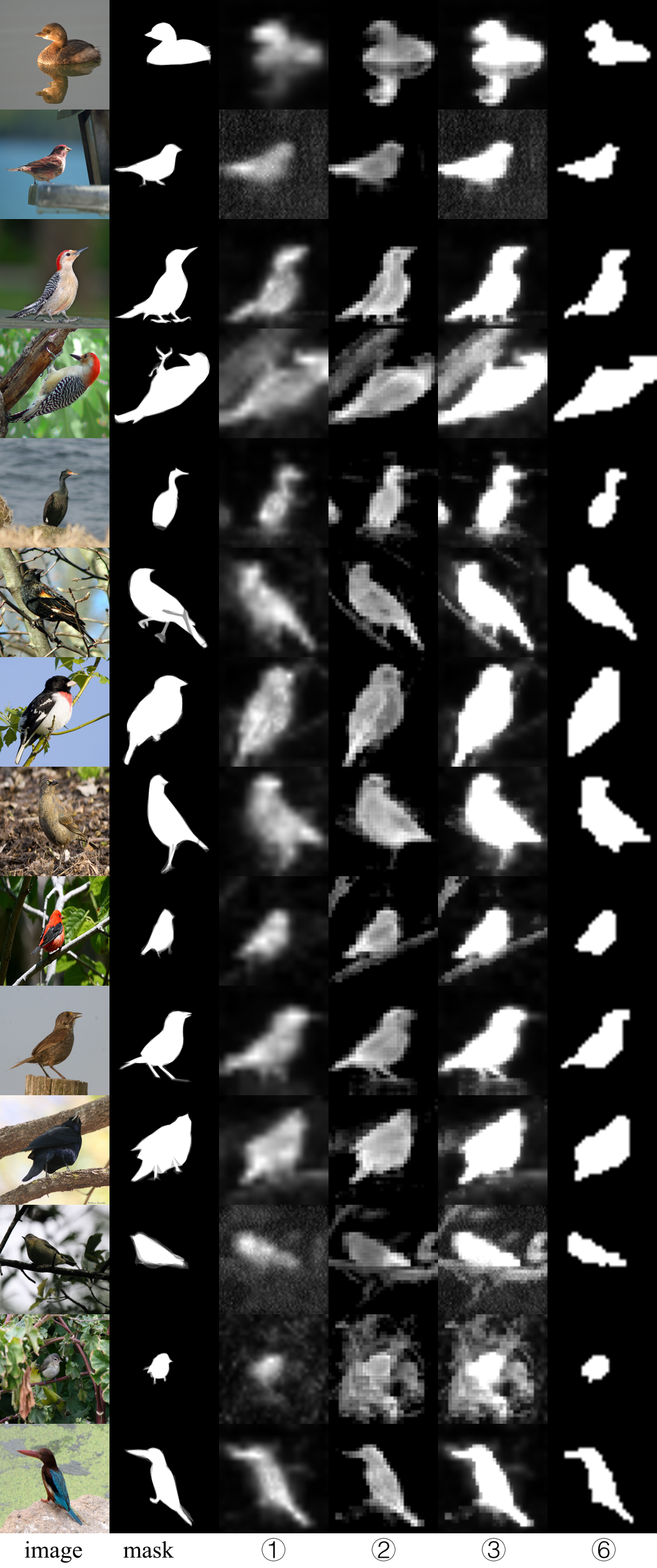}
    \caption{\textbf{Qualitative Results of AttentionCut on CUB.} 
    The columns from left to right are input image, ground-truth mask and \ding{172}, \ding{173}, \ding{174}, \ding{177} are the same as defined in \tabref{tab:attentioncut} (\ding{172}: only $\mathcal{A}_c$; \ding{173}: only $r(p)$; \ding{174}: $\mathcal{A}_c$ with $r(p)$, \ie, $\phi(p)$; \ding{177}: $\phi(p)$ with $\psi(p,q)$, \ie, AttentionCut).
    }
    \label{fig:cub}
\end{figure*}

\begin{figure*}[!ht]
    \centering
    \includegraphics[width=0.48\textwidth]{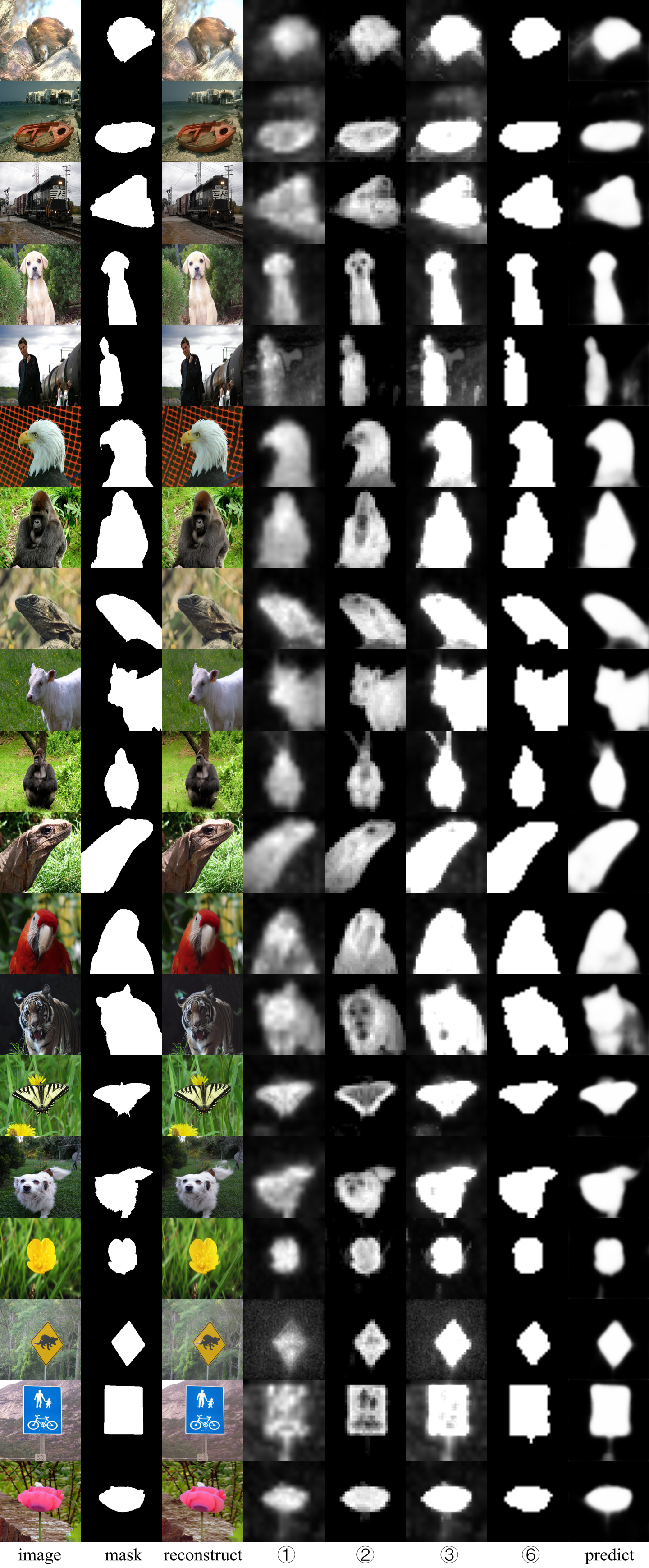}
    \includegraphics[width=0.48\textwidth]{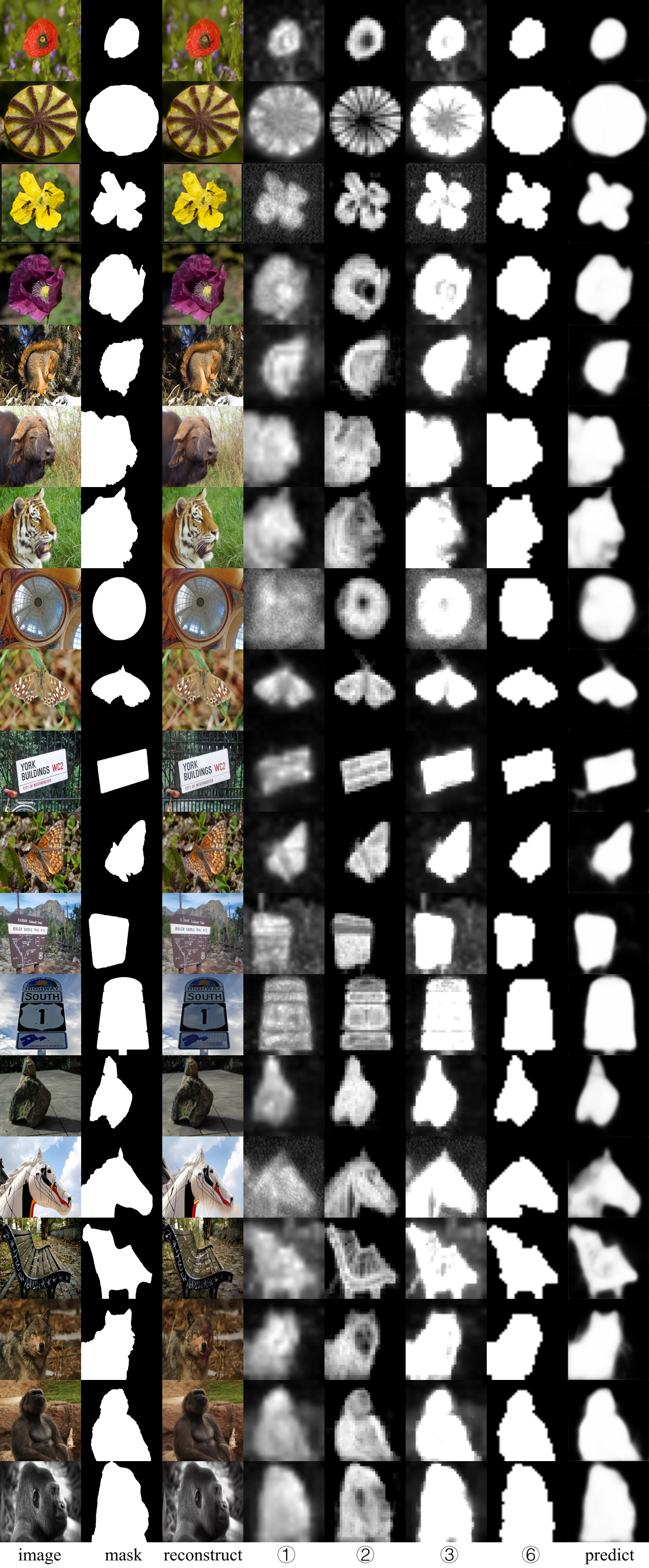}
    \caption{\textbf{Qualitative Results on ECSSD.} 
    First three columns are input image, ground-truth mask and reconstructed image by diffusion inversion.  
    \ding{172}, \ding{173}, \ding{174}, \ding{177} are the same as defined in \tabref{tab:attentioncut} (\ding{172}: only $\mathcal{A}_c$; \ding{173}: only $r(p)$; \ding{174}: $\mathcal{A}_c$ with $r(p)$, \ie, $\phi(p)$; \ding{177}: $\phi(p)$ with $\psi(p,q)$, \ie, AttentionCut). The last column is the prediction of \methodName.
    } 
    \label{fig:ecssd}
\end{figure*}

\begin{figure*}[!ht]
    \centering
    \includegraphics[width=0.48\textwidth]{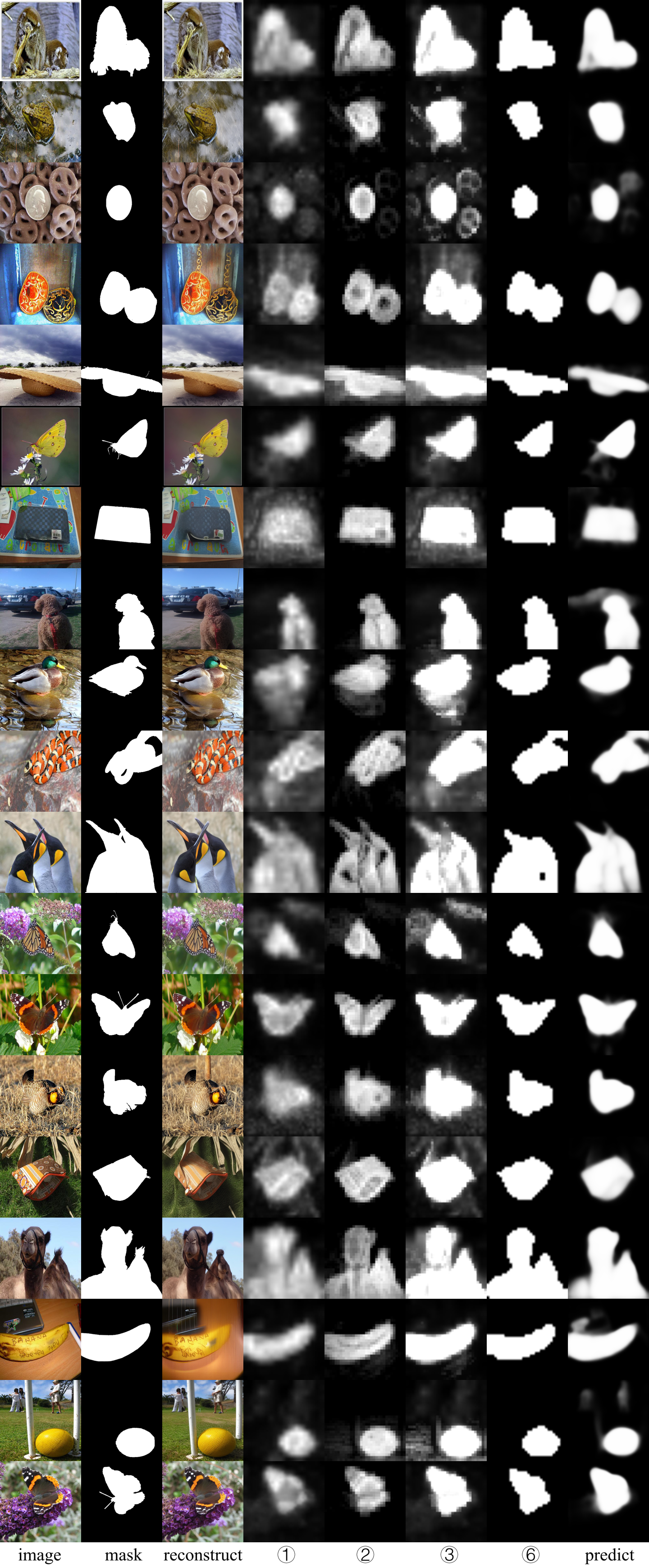}
    \includegraphics[width=0.48\textwidth]{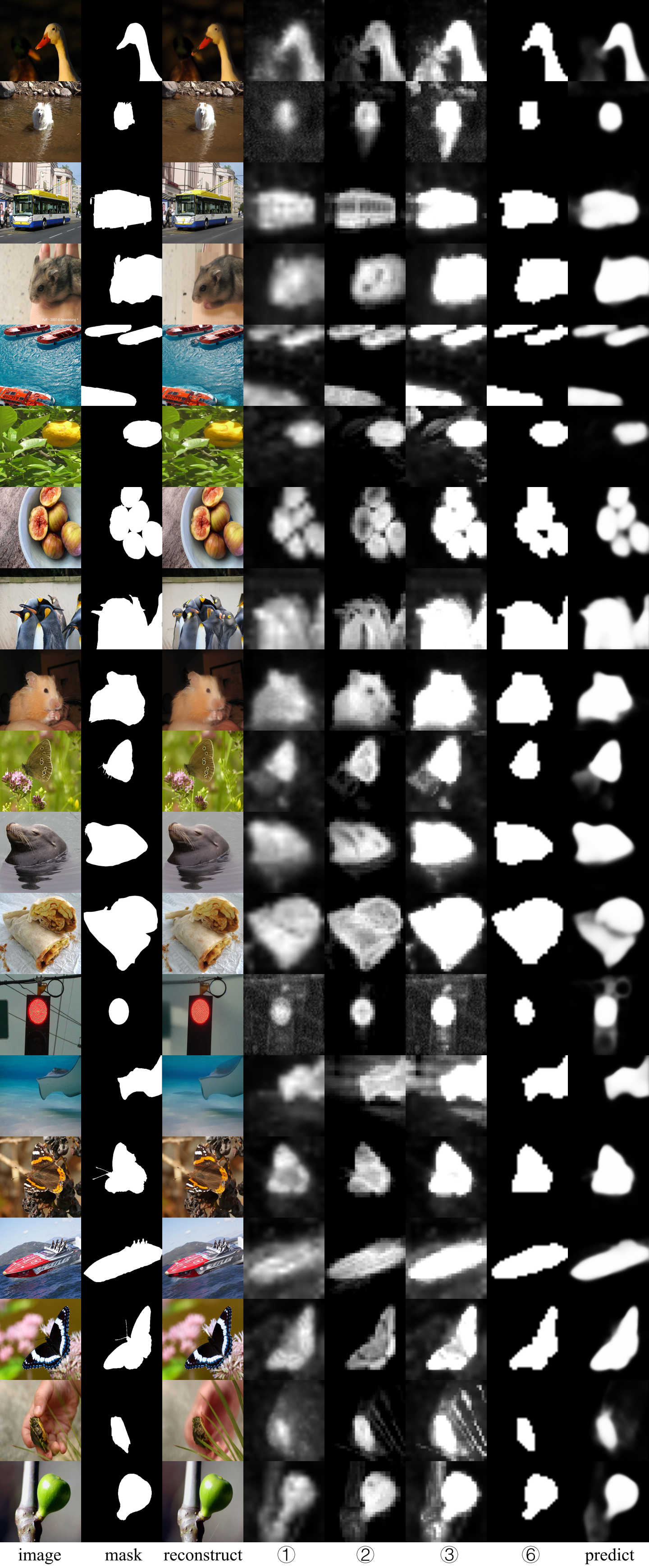}
    \caption{\textbf{Qualitative Results on DUTS-TE.} 
    First three columns are input image, ground-truth mask and reconstructed image by diffusion inversion.  
    \ding{172}, \ding{173}, \ding{174}, \ding{177} are the same as defined in \tabref{tab:attentioncut} (\ding{172}: only $\mathcal{A}_c$; \ding{173}: only $r(p)$; \ding{174}: $\mathcal{A}_c$ with $r(p)$, \ie, $\phi(p)$; \ding{177}: $\phi(p)$ with $\psi(p,q)$, \ie, AttentionCut). The last column is the prediction of \methodName.
    } 
    \label{fig:duts-te}
\end{figure*}

\begin{figure*}[!ht]
    \centering
    \includegraphics[width=0.48\textwidth]{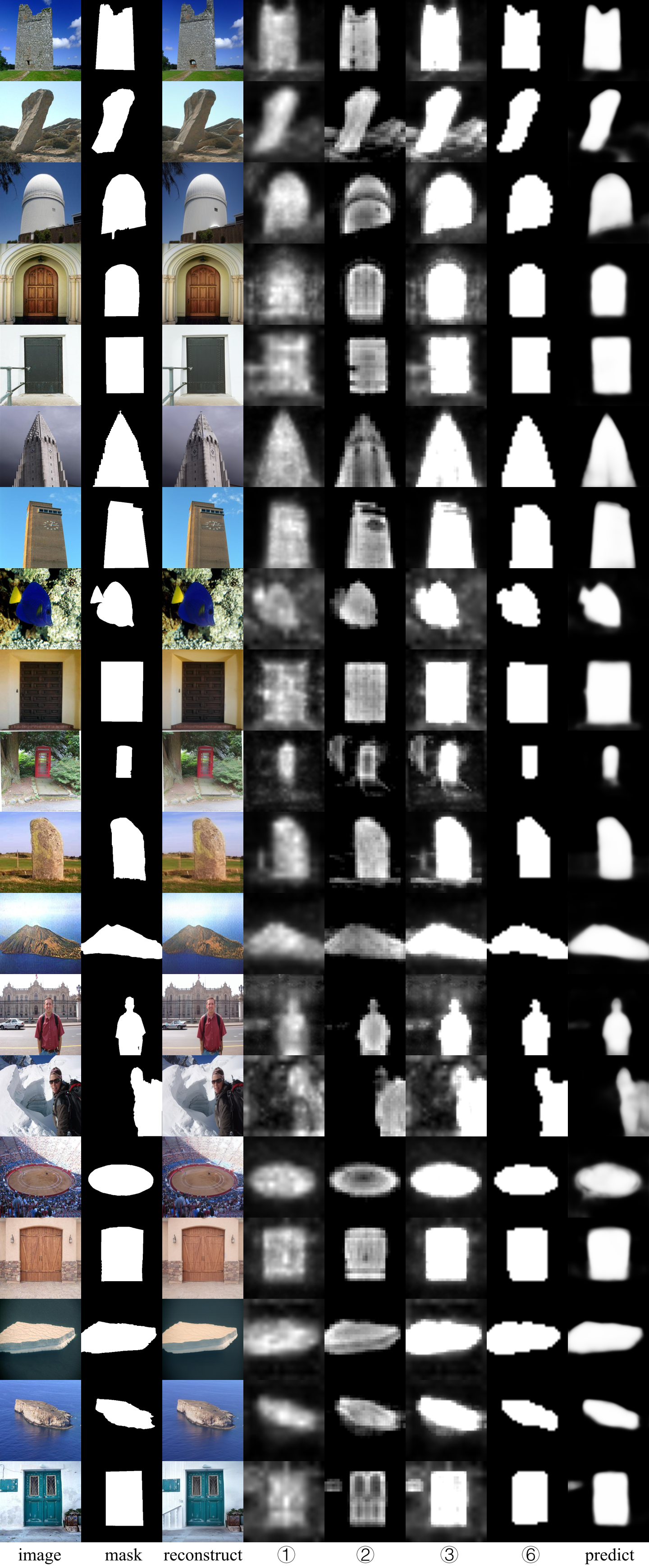}
    \includegraphics[width=0.48\textwidth]{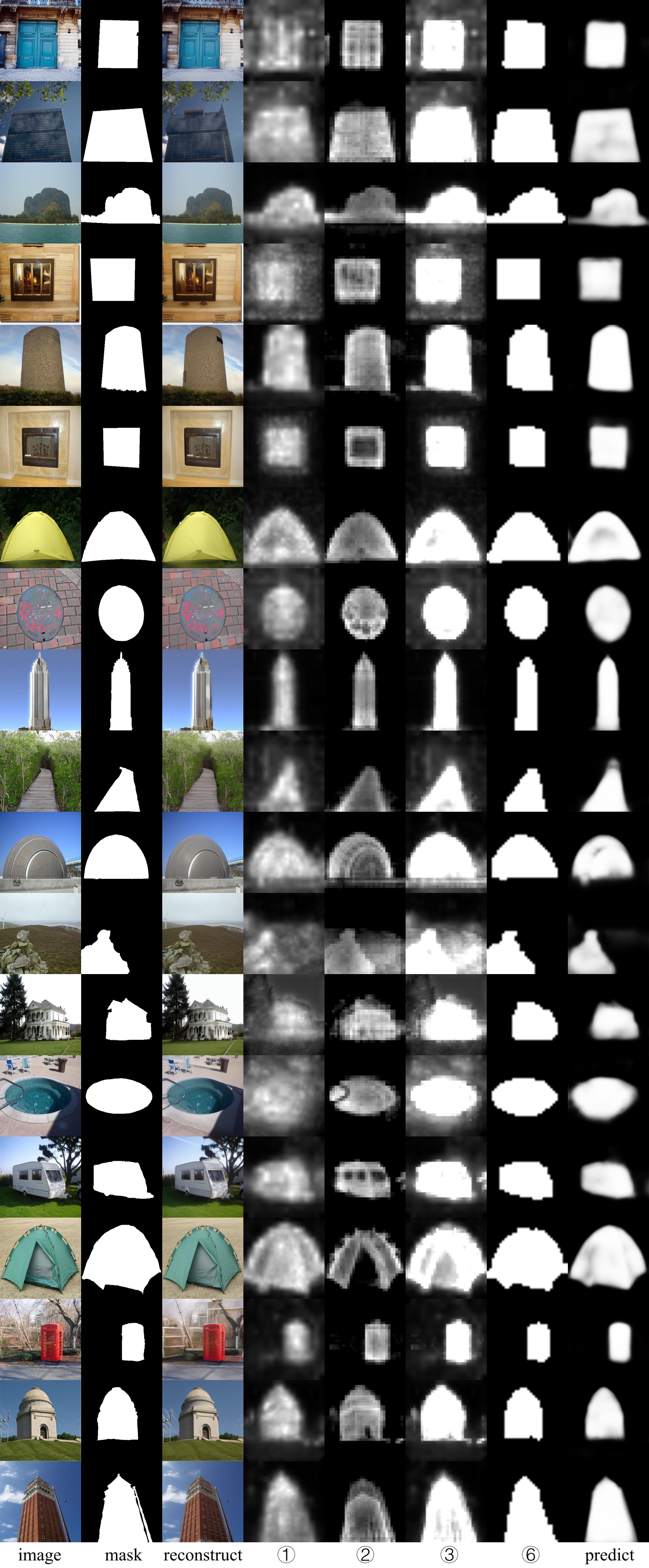}
    \caption{\textbf{Qualitative Results on DUT-OMRON.} 
    First three columns are input image, ground-truth mask and reconstructed image by diffusion inversion.  
    \ding{172}, \ding{173}, \ding{174}, \ding{177} are the same as defined in \tabref{tab:attentioncut} (\ding{172}: only $\mathcal{A}_c$; \ding{173}: only $r(p)$; \ding{174}: $\mathcal{A}_c$ with $r(p)$, \ie, $\phi(p)$; \ding{177}: $\phi(p)$ with $\psi(p,q)$, \ie, AttentionCut). The last column is the prediction of \methodName.
    } 
    \label{fig:duts-omron}
\end{figure*}

\begin{figure*}[!ht]
    \centering
    \includegraphics[width=0.98\textwidth]{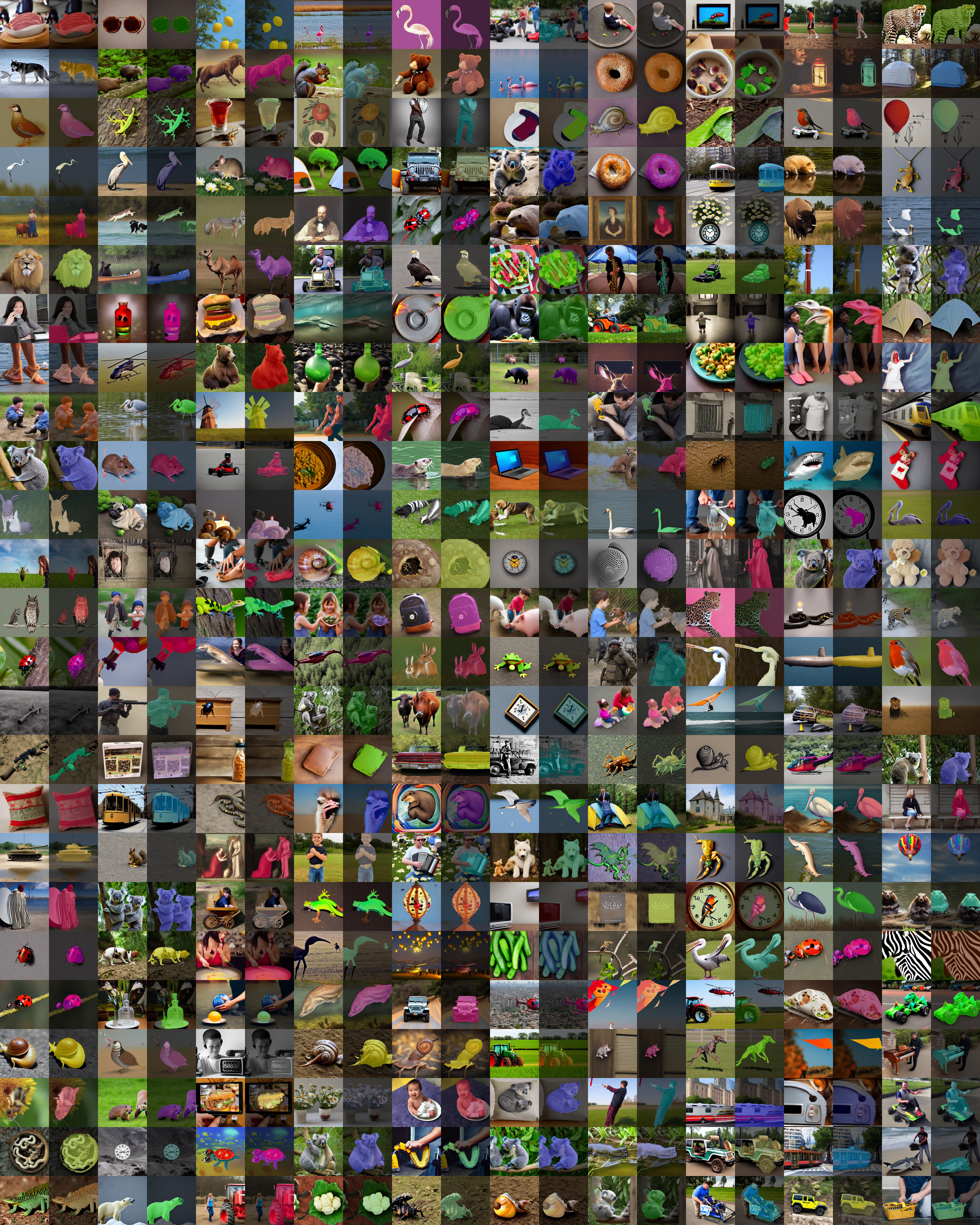}
    \caption{\textbf{Synthetic Image-mask Pairs.} With \textit{zero} human annotated required, \methodName can generate ``infinite'' realistic and diverse images together with impressive masks. Random samples are shown here.} 
    \label{fig:synthetic}
\end{figure*}

In this section, 
we first present qualitative visualizations of CUB~\cite{cub} on AttentionCut (\figref{fig:cub}), then visualizations of ECSSD~\cite{ecssd} (\figref{fig:ecssd}), DUTS-TE~\cite{duts} (\figref{fig:duts-te}), DUT-OMRON~\cite{dut_omron} (\figref{fig:duts-omron}) on both AttentionCut and \methodName, respectively. 
\figref{fig:synthetic} shows our synthetic dataset. 
Note that \ding{172}, \ding{173}, \ding{174}, \ding{177} are the same meaning as in \tabref{tab:attentioncut}. 
\ding{172}: only $\mathcal{A}_c$;
\ding{173}: only $r(p)$;
\ding{174}: $\mathcal{A}_c$ with $r(p)$, \ie, $\phi(p)$;
\ding{177}: $\phi(p)$ with $\psi(p,q)$, \ie, AttentionCut.

\twocolumn